\documentclass{article}

\PassOptionsToPackage{numbers,compress}{natbib}
\usepackage{amsmath} 
\usepackage{enumitem}
\usepackage{graphicx}
\usepackage{hyperref}       %
\usepackage{cleveref}
\usepackage{xcolor}
\usepackage{wrapfig}

\crefname{figure}{Fig.}{Figs.}  %
\Crefname{figure}{Fig.}{Figs.}  %

\crefname{table}{Tab.}{Tabs.}
\Crefname{table}{Tab.}{Tabs.}

\usepackage{graphicx} %

 \usepackage[main, final]{neurips_2026}

\usepackage[utf8]{inputenc} %
\usepackage[T1]{fontenc}    %
\usepackage{url}            %
\usepackage{booktabs}       %
\usepackage{amsfonts}       %
\usepackage{nicefrac}       %
\usepackage{microtype}      %
\usepackage{xcolor}         %

\usepackage{multirow}   %
\usepackage{amssymb}    %
\usepackage{graphicx}
\usepackage[table]{xcolor}
\usepackage[normalem]{ulem}

\title{Stable3R: Streaming 3D Reconstruction with Stable Geometric References}

\author{%
  Jincheng Xiong\textsuperscript{1, 2 *}, \hspace{0.75em}
  Jiarong Han\textsuperscript{2 *}, \hspace{0.75em}
  Hang Zhang\textsuperscript{2}, \hspace{0.75em}
  Mu Xu\textsuperscript{2}, \hspace{0.75em}
  Ming Qian\textsuperscript{2 \dag}\\
  \textsuperscript{1}Zhejiang University  \hspace{0.75em}  \textsuperscript{2}AMAP, Alibaba \\ [0.5em]
  \url{https://venite-xjc.github.io/Stable3R-page/}
}

\definecolor{linkblue}{RGB}{50,115,180}
\hypersetup{
    colorlinks=true,
    citecolor=linkblue, 
    linkcolor=linkblue, 
    urlcolor=linkblue  
}

\begin{document}

\renewcommand\thefootnote{}\footnotetext{$^*$ Equal contribution.}
\renewcommand\thefootnote{}\footnotetext{$^\dag$ Corresponding author.}

\maketitle

\begin{abstract}

Streaming 3D reconstruction has become a critical component for real-time applications. While recent advances have adapted large-scale offline feedforward models into streaming architectures via causal attention and KV-caching,  we found they suffer from fundamental instability: the irreversible bias in early frames. This limitation stems from two coupled factors: (1) the existing KV-cache-based causal architecture strictly prohibits access to future information, resulting in information-poor early caches; and (2) current methods typically anchor the pose constraints to the initial frame, amplifying the early bias throughout the sequence.
To address these challenges, we propose Stable3R, a holistic framework that establishes stable geometric references by {refining} historical caches and anchoring poses to prefixes rather than a single initial frame.
Specifically, we first introduce a lookahead-augmented attention mechanism, integrating future observations into past caches. {Second, we build on a frame-equivariant
architecture and adapt relative-pose supervision to causal prefixes, alleviating the reliance on the biased initial frame.}
Extensive experiments across multiple benchmarks show that our method consistently boosts reconstruction quality within a strictly causal inference schedule.

\end{abstract}

\section{Introduction}
\label{sec.intro}

Streaming 3D reconstruction is increasingly critical for real-time applications such as embodied robotics~\cite{wang2024embodiedscan} and interactive scene understanding~\cite{wu2025embodiedocc}. Existing approaches broadly fall into two categories. The first employs recurrent mechanisms~\cite{wang2025continuous, chen2025ttt3r}, compressing scenes into online updated states for continuous 3D perception, yet these methods often suffer from catastrophic forgetting. In contrast, causal streaming methods~\cite{streamVGGT, stream3r2025} extend offline 3D foundation models (e.g., VGGT~\cite{wang2025vggt}) to streaming settings via causal attention and KV-caches, enabling more continuous and stable retention of historical information and thereby improving reconstruction quality. However, we observe a key stability bottleneck in this paradigm: an over-reliance on early frames that contain limited observations.

\begin{figure}[t]
\centering
\includegraphics[width=\linewidth]{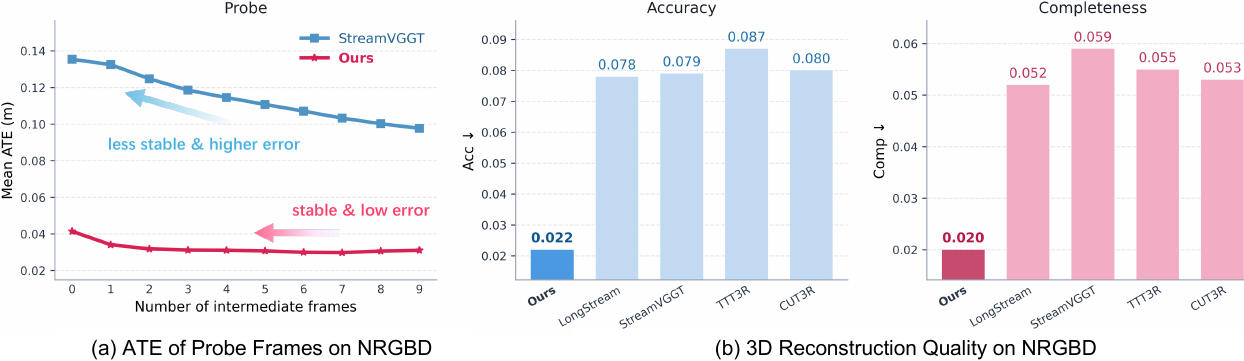}
\caption{Results on NRGBD~\cite{nrgbd} dataset. 
(a)
We introduce geometric context probes to evaluate the stability of geometric references in the KV caches. The protocol uniformly samples 30 images on each sequence, fixes the initial frame and final 20 probe frames while varying the number of intermediate frames, and measures the mean Absolute Trajectory Error (ATE) of the probe frames across sequences. We observe that the ATE of StreamVGGT~\cite{streamVGGT} varies significantly with the number of intermediate frames, indicating unstable geometric references in its historical observations. Moreover, earlier frames are inherently less reliable due to limited observations. In contrast, our method achieves lower ATE by stabilizing geometric references in the historical observations.
(b)
Our model achieves a more accurate 3D reconstruction than prior streaming methods.
}
\label{fig:teaser}
\end{figure}

The current causal streaming models~\cite{streamVGGT, stream3r2025} are inherently limited by two coupled factors: (i) the causal attention mechanism, which compels early frames to function on minimal geometric observations that cannot be retrospectively refined; and (ii) a "first-frame-anchored" architecture that ties global coordinate references and all pose regressions to the initial yet information-poor estimate. 
While first-frame-anchored designs are commonplace in offline models~\cite{wang2024dust3r, yang2025fast3r, wang2025vggt} with global information access, this architecture imposes a critical constraint in the streaming regime: the entire reconstruction becomes highly sensitive to the quality of the first frame. {Since standard causal attention prevents later observations from refining early representations}, all subsequent observations are forced to align with the frame containing the least amount of information. This mechanism inevitably leads to persistent instability. Our empirical analysis, utilizing geometric context probes as shown in ~\cref{fig:teaser} (a), confirms that reconstruction quality is positively correlated with the information density of early observations, demonstrating that stability is contingent upon the accumulation of rich historical context.

To overcome the aforementioned constraints, we propose Stable3R, a unified framework designed for stable streaming 3D reconstruction. We identify two key factors contributing to early-frame bias and mitigate their effects through complementary designs: First, we introduce \textbf{lookahead attention} to alleviate the information deficiency of causal {caches}, 
dynamically integrating future observations back into historical caches via a tailored parallel computation paradigm. Second, we adopt a \textbf{frame-equivariant architecture} where all frames share a unified token representation to eliminate structural asymmetry and replace absolute pose supervision with \textbf{prefix-relative pose supervision},
anchoring current estimates against preceding frames rather than a single initial noisy frame.
Consequently, our model exhibits enhanced stability against historical observations (see ~\cref{fig:teaser} (a)). Comprehensive evaluations on various benchmarks further demonstrate that our method achieves superior performance in both accurate camera pose estimation and robust 3D reconstruction across Sparse, Mid-length, and Long-sequence settings.

\section{Related Works}
\label{related_work}

\paragraph{Classical 3D Reconstruction.}
3D reconstruction, a fundamental problem in computer vision, aims to recover the geometric structure of objects or scenes from 2D observations.
Traditional approaches, such as Structure-from-Motion (SfM)~\cite{hartley2003multiple, schonberger2016structure} and Simultaneous Localization and Mapping (SLAM)  ~\cite{mur2015orb,mur2017orb,campos2021orb}, rely on hand-crafted features and complex pipelines. 
They typically involve feature extraction, matching, triangulation, and bundle adjustment 
to estimate camera poses and sparse keypoints, followed by Multi-View Stereo (MVS) methods~\cite{barnes2009patchmatch, bleyer2011patchmatch, galliani2015massively, schonberger2016pixelwise, sun2003stereo, nest-splatting, gaussian-surfel} to recover dense geometry. These pipelines are often time-consuming and suffer from limited robustness, particularly in texture-poor or challenging scenarios, lacking reliable visual features for tracking and reconstruction.

\paragraph{Offline 3D Reconstruction}
Data-driven transformer-based models have shifted 3D reconstruction toward a feed-forward paradigm. 
DUSt3R~\cite{wang2024dust3r} and its extension MASt3R~\cite{leroy2024grounding} directly regress 3D point maps from a pair of uncalibrated RGB images, and extend to multi-view settings via global alignment. Subsequent work ~\cite{yang2025fast3r, zhang2025flare} further generalizes this paradigm to multi-view inputs. Recently, VGGT~\cite{wang2025vggt} pioneers a simplified alternating attention architecture, demonstrating remarkable reconstruction quality via large-scale pretraining and multi-task learning. 
This has inspired a wave of feed-forward methods, ranging from adaptations for dynamic scenes~\cite{chen2025easi3r,wang20254d} to approaches enhancing accuracy and efficiency~\cite{wang2025pi,lin2025depth,jin2026zipmap,xie2026scal3r, liu2026planar}. However, these methods are inherently offline and computationally intensive, which precludes their use in real-time interactive applications.

\paragraph{Streaming 3D Reconstruction}
Unlike single-pass feed-forward models, streaming reconstruction enables continuous inference under partial observations, which is essential for real-time applications. Existing approaches are primarily built on two paradigms: recurrent/state-based models and causal streaming models. The former~\cite{wang2025continuous,chen2025ttt3r,wint3r} compress past observations into online states for efficient inference, but often suffer from forgetting or loss of fine-grained geometry. Other related directions include chunk-wise streaming designs such as LoGeR~\cite{zhang2026loger}, which are less aligned with strictly real-time causal inference, and hybrid methods~\cite{deng2025vggt,maggio2025vggt,maggio2026vggtslam20realtimedense}, which combine 3D foundation models with traditional SLAM pipelines but typically rely on hand-crafted components and system-level tuning. In contrast, causal streaming methods~\cite{streamVGGT,stream3r2025} extend offline 3D foundation models such as VGGT~\cite{wang2025vggt} to streaming settings via causal attention and KV-caches, achieving stronger reconstruction quality through direct retention of historical information. Subsequent train-free extensions~\cite{yuan2026infinitevggt,lu2026ovggt,xu2026framevggt} and concurrent long-horizon works, including LongStream~\cite{cheng2026longstream}, LingBot-Map~\cite{chen2026geometric} and ABot-Recon~\cite{abot-recon}, further improve scalability to longer sequences. However, existing causal streaming frameworks remain constrained by strict causality and first-frame dependence, which make reconstruction highly sensitive to early, information-poor observations. Our work instead focuses on addressing this stability bottleneck throughout the full streaming process.

\paragraph{Causal Attention in Streaming Models}

Causal attention is widely used in streaming models by masking future information, 
but it introduces an asymmetric information flow where early tokens cannot access future evidence, 
leading to limited global context. This issue has been observed in large language models~\cite{zhang2025diffusion,du2022glm,xiao2023efficient}, 
vision-language models~\cite{pei2025rethinking}, and world models~\cite{zhu2026causal,han2026grn}.
Prior work mitigates this by augmenting causal attention with additional context, 
including hybrid causal-bidirectional masks~\cite{boizard2026bidirlm,raffel2020exploring,du2022glm}, 
global signals such as contextual tokens~\cite{lin2025causal2vec} or decoder modifications~\cite{li2024bellm}, 
and retrieval or selection mechanisms~\cite{cai2025mixture,leviathan2024selective}.
The most related work, CASTLE~\cite{song2025causal}, introduces lookahead keys to inject future information into historical representations, partially alleviating the asymmetry caused by causal attention. This aligns with our observation that sparse early observations lead to biased scene estimation.
However, CASTLE assumes token-level causal attention and its parallelization scheme cannot be directly applied to our frame-level setting. We therefore implement dedicated backend kernels tailored to our lookahead attention.

\section{Method}

We propose a streaming 3D reconstruction framework that stabilizes early scene understanding through lookahead attention, while effectively mitigating early-frame bias via frame-equivariant modeling and prefix-relative pose supervision. In the following sections, we detail the streaming architecture, the lookahead attention mechanism, and the learning objectives.

\subsection{Streaming 3D Reconstruction}

\begin{figure}[ht]
\centering
\includegraphics[width=\linewidth]{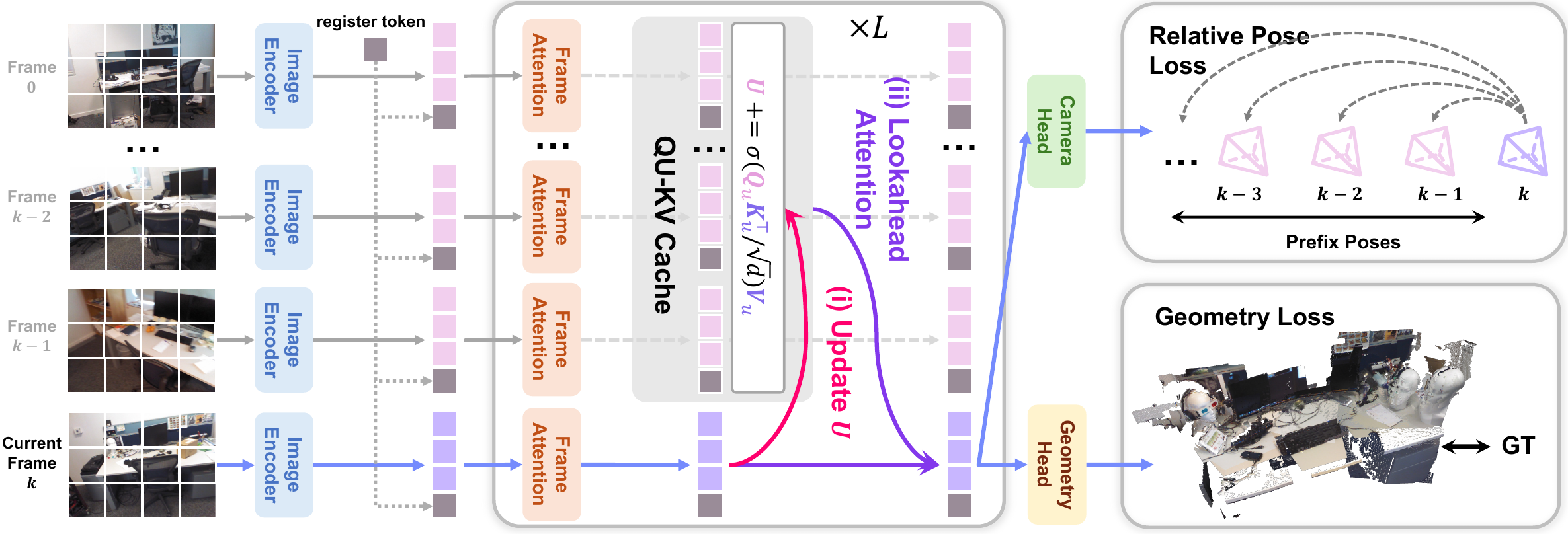}
\caption{
Overview of the proposed Stable3R framework for streaming 3D reconstruction. 
Pink denotes historical frames, while purple denotes the current frame.
We cache the causal states $(\boldsymbol{K}_c, \boldsymbol{V}_c)$ together with the lookahead states $(\boldsymbol{Q}_u, \boldsymbol{U})$. 
When a new frame arrives, the accumulated lookahead representation $\boldsymbol{U}$ is first updated using the incoming observations, followed by lookahead attention computation. 
In practice, both stages are fused into customized Triton kernels during training.  
}
\label{fig:pipeline}
\end{figure}

Given an input image sequence $\{I_i\}_{i=1}^N$, the model $\phi$ processes each frame $I_i$ sequentially and predicts the corresponding camera pose $\hat{T}_i \in \mathbb{R}^{4\times4}$ and point map $\hat{P}_i \in \mathbb{R}^{H\times W \times 3}$ using only the observed prefix $\{I_j\}_{j=1}^i$. Following prior streaming 3D reconstruction architectures~\cite{streamVGGT, stream3r2025}, we decompose $\phi$ into an encoder, a decoder, and a prediction head, as illustrated in \cref{fig:pipeline}:
\begin{equation}
F_i = \mathrm{Encoder}(I_i), \qquad
(G_i, \mathcal{M}_i) = \mathrm{Decoder}(F_i, \mathcal{M}_{i-1}), \qquad
\{\hat{T}_i, \hat{P}_i\} = \mathrm{Head}(G_i),
\end{equation}
where $\mathcal{M}_{i-1}$ denotes the cached historical memory constructed from the observed prefix $\{I_j\}_{j=1}^{i-1}$.

The encoder extracts per-frame visual tokens using a DINOv2-based tokenizer~\cite{oquab2023dinov2}. The decoder transforms these tokens into geometry-aware representations through alternating frame-wise attention and causal temporal attention. To reduce structural asymmetry across frames, we adopt a frame-equivariant architecture following $\pi^3$~\cite{wang2025pi}, where every frame is represented with the same token layout of patch tokens and register tokens. The decoder module propagates scene information across time through temporal attention with cached historical states. Finally, an MLP-based prediction head regresses the camera pose and point map for the current frame.

\subsection{Lookahead Attention}
\label{sec:lookahead_attention}

A critical challenge in streaming reconstruction is that early representations are derived from sparse observations, rendering them inherently noisy and incomplete. Under standard causal attention, historical memory is computed once and subsequently frozen, 
making early deviations difficult to rectify and adversely affecting later predictions. 
We overcome this limitation with lookahead attention, which iteratively refines the key-side representation of historical frames as new observations arrive, while remaining fully compatible with streaming inference: 
at time $i$, the model only accesses frames in $\{I_1,\dots,I_i\}$. 
Importantly, this refinement does not trigger retrospective prediction updates; instead, it is used only to improve subsequent predictions.

\textbf{Standard frame-level causal attention. }
We first recall standard frame-level causal attention in its parallel form. Given temporal frame tokens $\boldsymbol{X}\in\mathbb{R}^{NT\times d}$ derived from frames $\{I_1,\dots,I_N\}$, where $N$ is the number of frames, $T$ is the number of tokens per frame, and $d$ is the hidden dimension, we compute $\boldsymbol{Q}_c=\boldsymbol{X}\boldsymbol{W}_c^{Q},\quad\boldsymbol{K}_c=\boldsymbol{X}\boldsymbol{W}_c^{K},\quad\boldsymbol{V}_c=\boldsymbol{X}\boldsymbol{W}_c^{V} \in \mathbb{R}^{NT\times d_{\text{head}}}$. The causal attention output is then given by\begin{equation}\mathrm{Attention}(\boldsymbol{X})=\mathrm{softmax}\left(\frac{\boldsymbol{Q}_c\boldsymbol{K}_c^{\top}}{\sqrt{d_{\text{head}}}}+\boldsymbol{M}_{c}\right)\boldsymbol{V}_c,\end{equation}where the frame-wise causal mask $\boldsymbol{M}_c$ is defined as\begin{equation}\boldsymbol{M}_c(n,m)=\begin{cases}0, & \text{if } \text{frameIdx}(m) \leq \text{frameIdx}(n), \\-\infty, & \text{otherwise}.\end{cases}\end{equation}

\textbf{Frame-level lookahead attention. }
To incorporate future observations into historical memory, we introduce 
$\boldsymbol{Q}_u=\boldsymbol{X}\boldsymbol{W}_u^{Q},\quad\boldsymbol{K}_u=\boldsymbol{X}\boldsymbol{W}_u^{K},\quad\boldsymbol{V}_u=\boldsymbol{X}\boldsymbol{W}_u^{V}, \in \mathbb{R}^{NT\times d_{\text{head}}}$ similar to CASTLE~\cite{song2025causal} to compute time-varying lookahead keys $\boldsymbol{U}\in \mathbb{R}^{NT\times d_{\text{head}}}$. Specifically, when processing frame $i$, for historical token $m$ in frame $j (j<i)$, we aggregate future information from frame $j+1$ to frame $i$ into the lookahead key $\boldsymbol{U}^m \in \mathbb{R}^{d_\text{head}}$  as: \begin{equation}\boldsymbol{U}^{m}_{j \rightarrow i}=\sum_{n}\sigma\!\left(\frac{(\boldsymbol{Q}_u^{m})^\top \boldsymbol{K}_u^{n}}{\sqrt{d_{\text{head}}}}\right)\boldsymbol{V}_u^{n},\end{equation}where $\sigma(\cdot)$ denotes the sigmoid function and $j < \text{frameIdx}(n) \le i.$ 
We then incorporate the lookahead keys into causal attention by modifying the attention logits:\begin{equation}\label{equ:lookahead_attenion}\boldsymbol{O}=\mathrm{softmax}\left(\frac{\boldsymbol{Q}_c\boldsymbol{K}_c^{\top}}{\sqrt{d_{\text{head}}}}-\mathrm{SiLU}\!\left(\frac{\boldsymbol{Q}_c\boldsymbol{U}^{\top}}{\sqrt{d_{\text{head}}}}\right)+\boldsymbol{M}_{c}\right)\boldsymbol{V}_c,\end{equation}where $\boldsymbol{O}$ denotes the attention output. 
The lookahead term introduces an adaptive bias from enriched historical memory to improve geometric stability. 
Following CASTLE, we adopt SiLU~\cite{elfwing2018sigmoid} for smoother gating and better training stability. 
To support streaming inference, we cache both the lookahead states $(\boldsymbol{Q}_u, \boldsymbol{U})$ and the causal states $(\boldsymbol{K}_c, \boldsymbol{V}_c)$, referred to as the \textbf{QU-KV cache}. 
As illustrated in \cref{fig:pipeline}, we first update the lookahead keys $U$ of the history caches by new frame, then perform attention with updated $U$.
\begin{figure}[t]
\centering
\includegraphics[width=\linewidth]{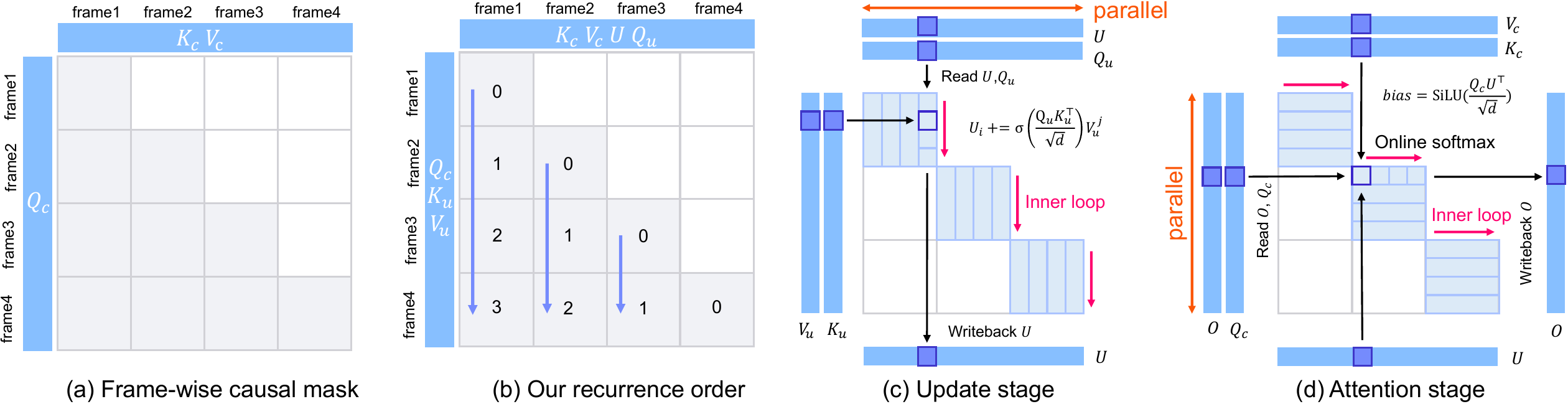}
\caption{
Efficient parallel lookahead attention under frame-causal (bidirectional + causal) attention. 
(a) Frame-wise causal masking induces a block-wise lower-triangular structure, enabling block-wise computation. 
(b) Blocks are processed in order of increasing diagonal offset (numbers indicate execution order). Blue arrows denote dependencies on the lookahead representation \(\boldsymbol{U}\). Each block is computed in two stages. 
(c) Update stage. \(\boldsymbol{Q}_u\) is tiled and processed in parallel, where each kernel instance scans \(\boldsymbol{K}_u, \boldsymbol{V}_u\) to accumulate \(\boldsymbol{U}\). 
(d) Attention stage. \(\boldsymbol{Q}_c\) is tiled and processed in parallel; each kernel performs online softmax over \(\boldsymbol{K}_c, \boldsymbol{V}_c\) with lookahead bias from \(\boldsymbol{U}\) to compute \(\boldsymbol{O}\).
}
\label{fig:flashattention}
\end{figure}

\textbf{Implementation for efficient parallel training.} 
Parallel computation is critical for training efficiency in streaming models. 
{CASTLE~\cite{song2025causal} proposes an efficient parallel training algorithm for lookahead attention based on strict token-level causality.
However, it is incompatible with our frame-causal attention, which is combined with bidirectional intra-frame and causal inter-frame interactions.}
To address this, we implement a dedicated parallelization scheme tailored to hybrid lookahead attention in Triton~\cite{tillet2019triton}.
Importantly, we observe that the inter-frame dependency remains causal, resulting in a block-wise lower-triangular structure (\cref{fig:flashattention}(a)) that admits block-wise decomposition~\cite{dao2022flashattention}.
For a historical frame \(j\), the lookahead keys progressively accumulate information from future frames. When interacting with a future frame \(i\), the lookahead keys \(\boldsymbol{U}_{j \rightarrow i}\) aggregate contributions from frames \(j+1\) to \(i\), yielding the following recursive update:
\begin{equation}
\label{equ:update_equation}
\boldsymbol{U}_{j \rightarrow i}
=
\boldsymbol{U}_{j \rightarrow i-1}
+
\sigma\!\left(
\frac{(\boldsymbol{Q}_u^{j})^\top \boldsymbol{K}_u^{i}}{\sqrt{d_{\text{head}}}}
\right)
\boldsymbol{V}_u^{i}.
\end{equation}
This recursive structure introduces dependencies across blocks along the frame dimension, 
which naturally leads to a diagonal scheduling order, as visualized in \cref{fig:flashattention}(b).

Concretely, we process the block matrix in increasing diagonal offsets, 
i.e., along diagonals with \(i - j = 0, 1, 2, \dots, N-1\), corresponding to the numbers in \cref{fig:flashattention}(b).
For each block \((j,i)\) in~\cref{fig:flashattention}(b), we perform two stages: 
(i) an update stage that performs \cref{equ:update_equation} to accumulate information from frame \(i\) into the lookahead keys of frame \(j\) , as shown in \cref{fig:flashattention}(c), and 
(ii) an attention stage that performs \cref{equ:lookahead_attenion} to compute lookahead attention from frame \(i\) to frame \(j\) using the updated lookahead keys, as shown in \cref{fig:flashattention}(d).
The update stage strictly follows a vertical dependency order, as indicated by the blue arrows in \cref{fig:flashattention}(b). Following FlashAttention-2~\cite{dao2023flashattention2}, we adopt query-aligned tiling in Triton~\cite{tillet2019triton} to reduce High Bandwidth Memory~(HBM)--Static Random-Access Memory~(SRAM) traffic. 
As the update and attention stages involve different queries \(\boldsymbol{Q}_u\) and \(\boldsymbol{Q}_c\), 
with distinct interaction directions, we decompose them into two kernels executed alternately, 
each parallelized along its query, leading to alternating tiling patterns.

For diagonal blocks \((j,j)\), no future information is involved, and thus the update stage is skipped. 
As a result, for \(N\) frames, the forward pass requires \(2N-1\) kernel launches. 
The backward pass follows the reverse (anti-diagonal) order to respect the dependency structure.

\subsection{Learning Objectives}

To mitigate early-frame bias in pose supervision, we adopt a prefix-relative pose objective. Instead of regressing each frame against a fixed global reference, we supervise relative camera poses over all causal pairs $(i,j)$ with $i<j$. This formulation has two advantages. First, it shifts supervision from absolute alignment to relative geometric consistency, making learning less sensitive to deviations in early frames.
Second, it provides a denser supervisory signal through the entire observed prefix, resulting in tighter constraints on the model compared to using a sparse initial reference.

Following MoGe~\cite{wang2025moge}, we first align the predicted and ground-truth pointcloud with a sequence-level scale factor $s^*$. Relative transformations are derived from predicted and ground-truth camera poses as
\begin{equation}
\hat{T}_{ij} = \hat{T}_i^{-1}\hat{T}_j, \qquad
T_{ij} = T_i^{-1}T_j,
\label{eq:relative_pose}
\end{equation}
with corresponding rotation and translation components $(\hat{R}_{ij}, \hat{t}_{ij})$ and $(R_{ij}, t_{ij})$. The pose loss is then defined as
\begin{equation}
\mathcal{L}_{\text{pose}}
=
\frac{1}{N_{\text{pair}}}
\sum_{i<j}
\left[
\lambda_r \mathcal{L}_{\text{rot}}(\hat{R}_{ij}, R_{ij})
+
\lambda_t \left\| s^* \hat{t}_{ij} - t_{ij} \right\|_{\text{Huber}}
\right],
\label{eq:pose_loss}
\end{equation}
where $N_{\text{pair}}$ is the number of causal frame pairs, $\mathcal{L}_{\text{rot}}$ denotes the rotation loss, and $\lambda_r,\lambda_t$ are balancing weights.

In addition, we adopt a scale-invariant local point-map loss following $\pi^3$~\cite{wang2025pi} to encourage accurate local geometry reconstruction:
\begin{equation}
\mathcal{L}_{\text{points}}
=
\frac{1}{3NHW}
\sum_{i=1}^{N}
\sum_{u=1}^{H \times W}
\frac{1}{z_{i,u}}
\left\|
s^* \hat{x}_{i,u} - x_{i,u}
\right\|_1,
\label{eq:points_loss}
\end{equation}
where $\hat{x}_{i,u}, x_{i,u} \in \mathbb{R}^3$ denote the predicted and ground-truth 3D coordinates at pixel $u$, and $z_{i,u}$ is the corresponding ground-truth depth used to normalize the residual.

The total training objective is a weighted sum of the pose and point-map losses:
\begin{equation}
\mathcal{L}_{\text{total}}
=
\lambda_{\text{pose}} \mathcal{L}_{\text{pose}}
+
\lambda_{\text{points}} \mathcal{L}_{\text{points}},
\end{equation}
where $\lambda_{\text{pose}}$ and $\lambda_{\text{points}}$ are balancing weights.

\section{Experiments}
\label{sec.experiments}

\subsection{Implementation Details} 
We train our model on a multi-domain dataset collection, including CO3Dv2~\cite{co3d}, WildRGBD~\cite{wildrgbd}, ARKitScenes~\cite{baruch2021arkitscenes}, ARKitScenes-HighRes~\cite{baruch2021arkitsceneshighres}, ScanNet++~\cite{yeshwanth2023scannet++}, ScanNet~\cite{dai2017scannet}, HyperSim~\cite{roberts2021hypersim}, BlendedMVS~\cite{yao2020blendedmvs}, MegaDepth~\cite{li2018megadepth}, Waymo~\cite{waymo}, VirtualKITTI2~\cite{vkitt2}, TartanAir~\cite{wang2020tartanair}, DL3DV~\cite{ling2024dl3dv}, OmniObject3D~\cite{wu2023omniobject3d}, Spring~\cite{mehl2023spring}, PointOdyssey~\cite{zheng2023pointodyssey}, UASOL~\cite{bauer2019uasol} and Matterport3D~\cite{map3d} datasets.

Initialized from pre-trained $\pi^3$ weights~\cite{wang2025pi}, our framework keeps the backbone encoder frozen. The proposed lookahead module is trained from scratch with a peak learning rate of $1 \times 10^{-4} $ , whereas the rest of the model uses a peak learning rate of $1 \times 10^{-5} $ . We employ the AdamW optimizer~\cite{adaw} for 24k steps, training on image sequences of 4 to 24 frames. The entire training process takes approximately 2 days on 32 NVIDIA H20 GPUs. 

\subsection{Experimental Setup} 

We evaluate Stable3R on two core tasks: 3D reconstruction and camera pose estimation. For 3D reconstruction, we consider three evaluation regimes: a \textbf{Sparse} setting with minimal or no frame overlap (2--5 frames), a \textbf{Mid-length} setting with uniformly sampled 25-frame sequences, and a \textbf{Long-sequence} setting with 300-frame sequences. We first report reconstruction results under the Sparse and Mid-length settings, then evaluate long-sequence generalization, and finally assess trajectory accuracy for camera pose estimation.

\subsection{3D Reconstruction Results} 
\subsubsection{Sparse and Mid-length Evaluation}

Following StreamVGGT and CUT3R~\cite{streamVGGT,wang2025continuous}, we evaluate reconstruction quality on both static scenes (NRGBD~\cite{nrgbd} and 7-Scenes~\cite{7scene}) and the more challenging dynamic benchmark TUM-dynamics~\cite{tum}. We report accuracy (Acc), completeness (Comp), and normal consistency (NC). Under the \textbf{Sparse} setting, we follow the StreamVGGT protocol to sample 2--4 frames on NRGBD and 3--5 frames on 7-Scenes, while extending the evaluation to TUM-dynamics with uniformly sampled 5 frames. Under the \textbf{Mid-length} setting, we uniformly sample 25 frames for each sequence.

We compare against strictly streaming baselines, including CUT3R~\cite{wang2025continuous}, TTT3R~\cite{chen2025ttt3r}, StreamVGGT~\cite{streamVGGT}, and LongStream~\cite{cheng2026longstream}. Following the CUT3R evaluation framework~\cite{wang2025continuous}, we benchmark all methods using their publicly available weights under a unified protocol.

\begin{figure}[t]
\centering
\includegraphics[width=\linewidth]{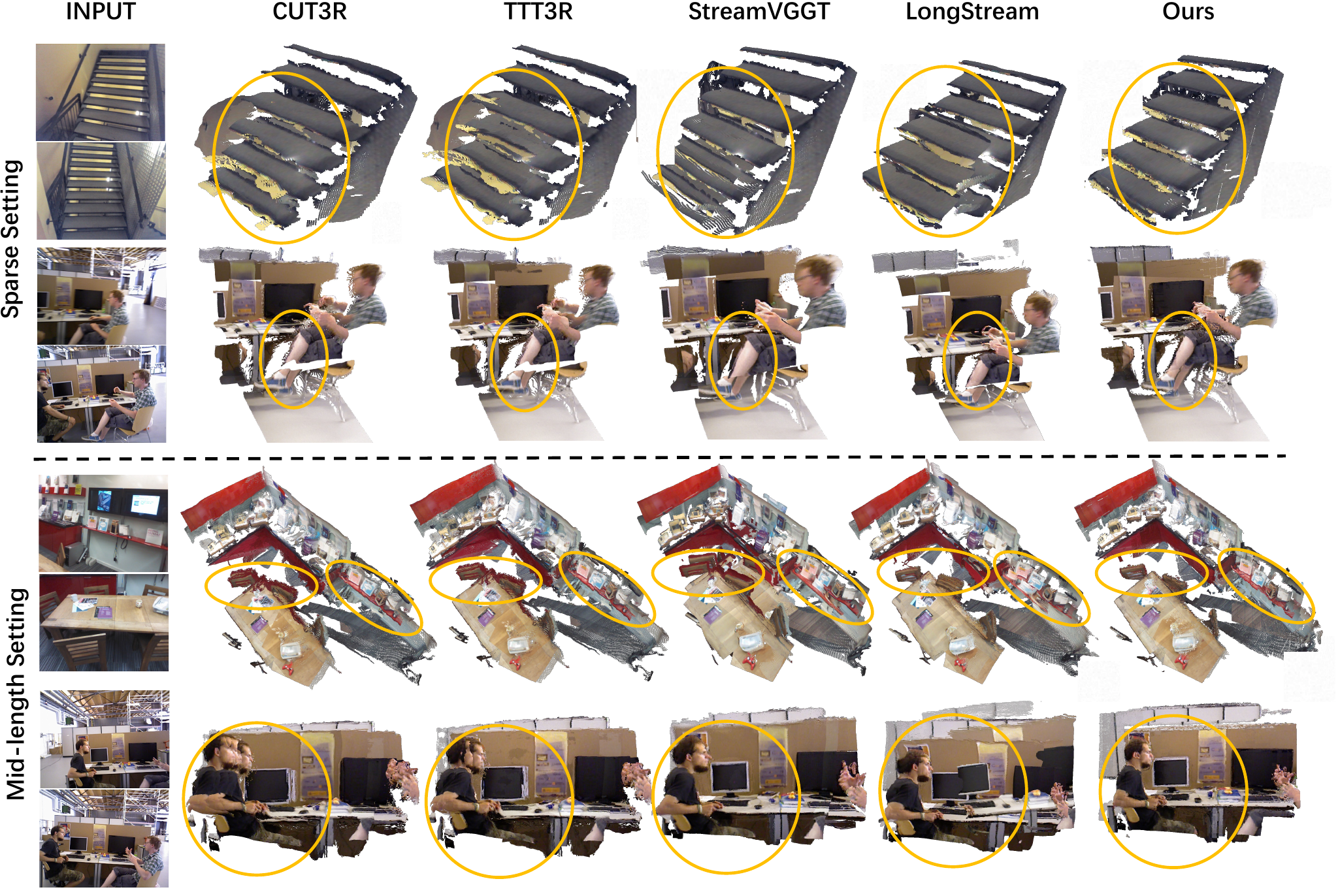}
\caption{
Qualitative Results of 3D Reconstruction.
}
\label{fig:expetiment1}
\end{figure}

As shown in \cref{tab:short_eval,fig:expetiment1}, Stable3R consistently improves reconstruction quality under both settings. In particular, the gains in accuracy and completeness are consistent with the visual results, where our method produces more coherent structures and significantly fewer artifacts. These results suggest that lookahead attention and frame-equivariant modeling with relative supervision stabilize the streaming process by mitigating the information deficiency of causal memory. Notably, the method remains robust on dynamic scenes (e.g., the bottom row of both settings in \cref{fig:expetiment1}), highlighting its potential for streaming 4D reconstruction in dynamic environments.

{To further evaluate the generalization ability of Stable3R, we conduct experiments on the recently introduced SpatialBench~\cite{peng2026spatialbench}, a cross-paradigm and domain-diverse benchmark, comparing against other streaming 3D reconstruction methods. We evaluate the sparse and medium settings, additionally reporting AbsRel for depth, AUC@30 for camera pose, and F-score for point clouds. As shown in~\cref{tab:spatialbench}, Stable3R achieves the best depth and camera pose results among streaming baselines under the sparse setting. Under the medium setting, Stable3R achieves the best results in depth, camera pose, and point cloud reconstruction, while ranking second in trajectory ATE, behind only LongStream. These results demonstrate the strong generalization ability of Stable3R across diverse domains.}

\begin{table}[htbp]
    \centering
    \setlength{\tabcolsep}{3.5pt}
    \renewcommand{\arraystretch}{1.08}
\caption{Quantitative 3D reconstruction results on NRGBD, 7-Scenes, and TUM-dynamics under Sparse and Mid-length settings. Bold and underlined values indicate the best and second-best results, respectively.}
    \label{tab:short_eval}
    
    \resizebox{\textwidth}{!}{
        \begin{tabular}{c l *{18}{c}}
            \toprule
            \multirow{3}{*}{} & \multirow{3}{*}{\textbf{Method}} 
            & \multicolumn{6}{c}{\textbf{NRGBD}} 
            & \multicolumn{6}{c}{\textbf{7-Scenes}} 
            & \multicolumn{6}{c}{\textbf{TUM-dynamics}} \\
            
            \cmidrule(lr){3-8} \cmidrule(lr){9-14} \cmidrule(lr){15-20}
            
            & & \multicolumn{2}{c}{\textbf{Acc}$\downarrow$} 
              & \multicolumn{2}{c}{\textbf{Comp}$\downarrow$} 
              & \multicolumn{2}{c}{\textbf{NC}$\uparrow$} 
              & \multicolumn{2}{c}{\textbf{Acc}$\downarrow$} 
              & \multicolumn{2}{c}{\textbf{Comp}$\downarrow$} 
              & \multicolumn{2}{c}{\textbf{NC}$\uparrow$} 
              & \multicolumn{2}{c}{\textbf{Acc}$\downarrow$} 
              & \multicolumn{2}{c}{\textbf{Comp}$\downarrow$} 
              & \multicolumn{2}{c}{\textbf{NC}$\uparrow$} \\
            
            \cmidrule(lr){3-4} \cmidrule(lr){5-6} \cmidrule(lr){7-8}
            \cmidrule(lr){9-10} \cmidrule(lr){11-12} \cmidrule(lr){13-14}
            \cmidrule(lr){15-16} \cmidrule(lr){17-18} \cmidrule(lr){19-20}
            
            & & \textbf{Mean} & \textbf{Med.} 
              & \textbf{Mean} & \textbf{Med.} 
              & \textbf{Mean} & \textbf{Med.} 
              & \textbf{Mean} & \textbf{Med.} 
              & \textbf{Mean} & \textbf{Med.} 
              & \textbf{Mean} & \textbf{Med.} 
              & \textbf{Mean} & \textbf{Med.} 
              & \textbf{Mean} & \textbf{Med.} 
              & \textbf{Mean} & \textbf{Med.} \\
            \midrule
            
            \multirow{5}{*}{\rotatebox[origin=c]{90}{\textbf{Sparse}}}
            & CUT3R~\cite{wang2025continuous} 
                & 0.096 & 0.048 & 0.084 & 0.043 & 0.822 & 0.963
                & 0.084 & 0.056 & 0.097 & 0.056 & 0.698 & 0.787
                & 0.148 & 0.099 & 0.095 & 0.049 & 0.643 & 0.718 \\
            
            & TTT3R~\cite{chen2025ttt3r} 
                & 0.100 & 0.055 & 0.085 & 0.049 & 0.820 & 0.960
                & 0.088 & 0.059 & 0.125 & 0.077 & 0.694 & 0.783
                & 0.148 & 0.096 & 0.096 & 0.050 & 0.646 & 0.721 \\

            & StreamVGGT~\cite{streamVGGT}
                & \underline{0.085} & \underline{0.044} & \underline{0.079} & \underline{0.038} & \underline{0.862} & \underline{0.986}
                & 0.132 & 0.058 & 0.116 & \underline{0.042} & \textbf{0.749} & \textbf{0.863}
                & 0.126 & 0.060 & 0.070 & 0.036 & \textbf{0.714} & \textbf{0.820} \\

            & LongStream~\cite{cheng2026longstream}
                & 0.091 & 0.063 & 0.106 & 0.078 & 0.814 & 0.956
                & \underline{0.068} & \underline{0.045} & \underline{0.069} & \textbf{0.041} & 0.738 & 0.843
                & \underline{0.080} & \textbf{0.032} & \underline{0.060} & \underline{0.031} & \underline{0.709} & \underline{0.816} \\

            & {Ours}
                & \textbf{0.035} & \textbf{0.020} & \textbf{0.041} & \textbf{0.023} & \textbf{0.872} & \textbf{0.987}
                & \textbf{0.052} & \textbf{0.032} & \textbf{0.069} & 0.048 & \underline{0.743} & \underline{0.847}
                & \textbf{0.079} & \underline{0.045} & \textbf{0.051} & \textbf{0.028} & 0.707 & 0.805 \\
                
            \midrule
            
            \multirow{5}{*}{\rotatebox[origin=c]{90}{\textbf{Mid-len.}}}
            & CUT3R~\cite{wang2025continuous} 
                & 0.080 & \underline{0.046} & 0.053 & 0.025 & 0.820 & 0.958
                & \underline{0.025} & \underline{0.013} & 0.032 & \underline{0.012} & 0.702 & 0.807
                & 0.106 & 0.049 & 0.057 & 0.020 & 0.577 & 0.615 \\

            & TTT3R~\cite{chen2025ttt3r} 
                & 0.087 & 0.049 & 0.055 & 0.028 & 0.808 & 0.948
                & 0.040 & 0.021 & 0.038 & 0.013 & 0.687 & 0.789
                & 0.102 & 0.049 & 0.051 & 0.019 & 0.580 & 0.620 \\

            & StreamVGGT~\cite{streamVGGT} 
                & 0.079 & 0.055 & 0.059 & 0.032 & \underline{0.846} & \underline{0.983}
                & 0.045 & 0.026 & 0.035 & 0.016 & 0.716 & 0.826
                & 0.091 & 0.046 & \underline{0.042} & 0.014 & \textbf{0.633} & \textbf{0.711} \\

            & LongStream~\cite{cheng2026longstream}
                & \underline{0.078} & 0.050 & \underline{0.052} & \underline{0.024} & 0.836 & 0.975
                & 0.030 & 0.019 & \underline{0.027} & 0.013 & \underline{0.720} & \underline{0.832}
                & \underline{0.072} & \underline{0.024} & 0.045 & \underline{0.012} & 0.615 & 0.682 \\

            & {Ours}
                & \textbf{0.022} & \textbf{0.014} & \textbf{0.020} & \textbf{0.010} & \textbf{0.884} & \textbf{0.987}
                & \textbf{0.019} & \textbf{0.010} & \textbf{0.022} & \textbf{0.010} & \textbf{0.722} & \textbf{0.834}
                & \textbf{0.048} & \textbf{0.015} & \textbf{0.031} & \textbf{0.011} & \underline{0.626} & \underline{0.697} \\
            
            \bottomrule
        \end{tabular}
    }
\end{table}

\begin{table}[htbp]
    \centering
    \caption{Comparison on the Sparse and Medium tasks of SpatialBench~\cite{peng2026spatialbench}. Best and second-best results are in \textbf{bold} and \underline{underlined}, respectively.}
    \label{tab:spatialbench}
    \resizebox{0.8\textwidth}{!}
    {
    \begin{tabular}{l cc cccc}
        \toprule
        \multirow{2}{*}{\textbf{Method}} & \multicolumn{2}{c}{\textbf{Sparse}} & \multicolumn{4}{c}{\textbf{Medium}} \\
        \cmidrule(lr){2-3} \cmidrule(lr){4-7}
         & \textbf{AbsRel $\downarrow$} & \textbf{AUC@30 $\uparrow$} & \textbf{AbsRel $\downarrow$} & \textbf{AUC@30 $\uparrow$} & \textbf{ATE $\downarrow$} & \textbf{F-Score $\uparrow$} \\
        \midrule
        CUT3R & 0.196 & 0.519 & 0.189 & 0.469 & 2.676 & 0.286 \\
        TTT3R & 0.202 & 0.469 & 0.179 & 0.493 & 2.343 & 0.294 \\
        StreamVGGT & 0.154 & \underline{0.598} & 0.171 & 0.562 & 4.940 & 0.397 \\
        InfiniteVGGT & 0.154 & 0.596 & 0.170 & \underline{0.563} & 4.964 & \underline{0.402} \\
        LongStream & \underline{0.151} & 0.543 & \underline{0.166} & 0.385 & \textbf{1.188} & 0.126 \\
        \rowcolor{gray!10}
        \textbf{Ours} & \textbf{0.105} & \textbf{0.628} & \textbf{0.097} & \textbf{0.642} & \underline{2.027} & \textbf{0.541} \\
        \bottomrule
    \end{tabular}
    }
\end{table}

\subsubsection{Long-Sequence Evaluation}

Although trained only on short sequences (up to 24 frames), Stable3R generalizes effectively to long sequences. We evaluate long-sequence extrapolation {under two inference settings: (i) direct inference with the original models and (ii) inference using} FrameVGGT's train-free, frame-level bounded-memory strategy~\cite{xu2026framevggt}. {For the latter setting, since the original implementation of FrameVGGT is unavailable}, we implement a closely related version based on the paper and apply it to both our method {(denoted as Ours*)} and StreamVGGT~\cite{streamVGGT} (denoted as FrameVGGT*). Concretely, we use a 16-frame mid-memory budget together with 4-frame sparse anchors.

We compare against state-memory baselines, including CUT3R~\cite{wang2025continuous} and TTT3R~\cite{chen2025ttt3r}, as well as KV-cache-based methods, including InfiniteVGGT~\cite{yuan2026infinitevggt}, OVGGT~\cite{lu2026ovggt}, and LongStream~\cite{cheng2026longstream}. Experiments are conducted on 7-Scenes~\cite{7scene} and NRGBD~\cite{nrgbd} using 300-frame samples.

{As shown in~\cref{tab:long_seq_results}, both inference variants of our method achieve strong geometric reconstruction performance across datasets while maintaining competitive normal consistency. In particular, our standard variant achieves the best reconstruction accuracy on both datasets, while the bounded-memory variant attains the best mean completeness on 7-Scenes. These results demonstrate strong generalization from short-sequence training to long-horizon inference, as well as compatibility with test-time optimization. Notably, LingBot-Map is specifically trained on sequences exceeding 300 frames with substantially greater training cost than ours.}

\begin{table}[htbp]
    \centering
    \caption{Quantitative 3D reconstruction results on 7-Scenes and NRGBD on long sequences.}
    \label{tab:long_seq_results}
    \resizebox{\textwidth}{!}{
    \begin{tabular}{l c c c c c c c c c c c c}
        \toprule
        \multirow{3}{*}{\textbf{Method}} & \multicolumn{6}{c}{\textbf{7-Scenes}} & \multicolumn{6}{c}{\textbf{NRGBD}} \\
        \cmidrule(lr){2-7} \cmidrule(lr){8-13}
        & \multicolumn{2}{c}{\textbf{Acc}$\downarrow$} & \multicolumn{2}{c}{\textbf{Comp}$\downarrow$} & \multicolumn{2}{c}{\textbf{NC}$\uparrow$} & \multicolumn{2}{c}{\textbf{Acc}$\downarrow$} & \multicolumn{2}{c}{\textbf{Comp}$\downarrow$} & \multicolumn{2}{c}{\textbf{NC}$\uparrow$} \\
        \cmidrule(lr){2-3} \cmidrule(lr){4-5} \cmidrule(lr){6-7} \cmidrule(lr){8-9} \cmidrule(lr){10-11} \cmidrule(lr){12-13}
        & \textbf{Mean} & \textbf{Med.} & \textbf{Mean} & \textbf{Med.} & \textbf{Mean} & \textbf{Med.} & \textbf{Mean} & \textbf{Med.} & \textbf{Mean} & \textbf{Med.} & \textbf{Mean} & \textbf{Med.} \\
        \midrule
        CUT3R~\cite{wang2025continuous} & 0.107 & 0.072 & 0.061 & 0.014 & 0.543 & 0.563 & 0.139 & 0.100 & 0.097 & 0.036 & 0.580 & 0.626 \\
        TTT3R~\cite{chen2025ttt3r} & 0.035 & 0.021 & 0.023 & \underline{0.004} & 0.567 & 0.603 & 0.100 & 0.056 & 0.048 & 0.024 & 0.604 & 0.670 \\
        LongStream~\cite{cheng2026longstream} & 0.077 & 0.047 & 0.030 & 0.006 & 0.568 & 0.605 & 0.119 & 0.082 & 0.028 & 0.006 & 0.630 & 0.726 \\
        InfiniteVGGT~\cite{yuan2026infinitevggt} & 0.041 & 0.021 & 0.026 & 0.005 & 0.565 & 0.599 & 0.067 & 0.044 & 0.022 & 0.004 & 0.634 & 0.730 \\
        OVGGT~\cite{lu2026ovggt} & 0.028 & 0.012 & 0.020 & \underline{0.004} & \underline{0.576} & \underline{0.617} & 0.033 & 0.023 & 0.011 & \underline{0.003} & 0.641 & \underline{0.738} \\
        FrameVGGT* & 0.029 & 0.013 & 0.022 & \underline{0.004} & \textbf{0.581} & \textbf{0.626} & 0.030 & 0.021 & 0.013 & 0.004 & \textbf{0.680} & \textbf{0.807} \\
        {LingBot-Map} & \underline{0.019} & \underline{0.009} & 0.018 & \textbf{0.003} & 0.571 & 0.609 & \underline{0.025} & \underline{0.016} & \textbf{0.007} & \textbf{0.002} & 0.618 & 0.693\\
        \rowcolor{gray!10}
        \textbf{Ours*} & 0.020 & 0.010 & \textbf{0.014} & \textbf{0.003} & 0.572 & 0.610 & 0.029 & 0.018 & \underline{0.009} & \underline{0.003} & 0.627 & 0.707 \\
        \rowcolor{gray!10}
        \textbf{Ours} 
        & \textbf{0.013} &\textbf{0.005} &\underline{0.015}& \underline{0.004}& 0.572 &0.611& \textbf{0.018}& \textbf{0.011}& \textbf{0.007} &\textbf{0.002} &\underline{0.645} &\underline{0.738}\\
        \bottomrule
    \end{tabular}
    }
\end{table}

\subsection{Camera Pose Estimation Results}

\begin{figure}[t]
\centering
\includegraphics[width=\linewidth]{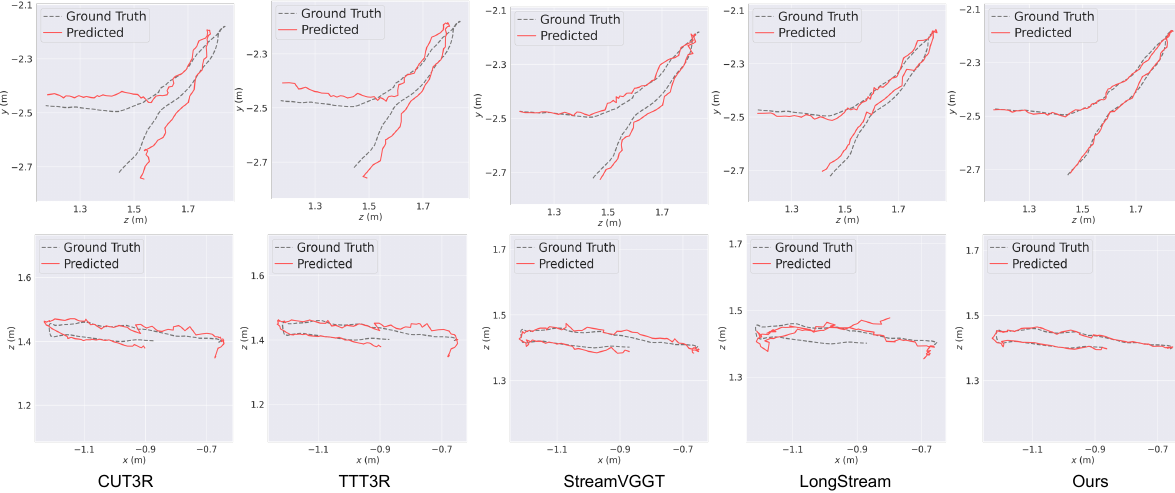}
\caption{
Pose visualization comparison on the TUM dataset.
}
\label{fig:expetiment2}
\end{figure}

We evaluate camera pose estimation on TUM-dynamics~\cite{tum} (90-frame sequences), Sintel~\cite{butler2012naturalistic} (20–50 frame sequences) and ScanNet~\cite{dai2017scannet} (90-frame sequences), following the StreamVGGT protocol. Aligned to ground truth via Sim(3), predictions are assessed using Absolute Translation Error (ATE), Relative Translation Error (RPE trans), and Relative Rotation Error (RPE rot). 
Notably, Sintel and TUM-dynamics contain dynamic objects, presenting significant challenges for geometric consistency in pose estimation.
 As shown in ~\cref{tab:pose_evaluation} and ~\cref{fig:expetiment2}, by providing more stable geometric references, our approach ensures superior trajectory accuracy in streaming scenarios.

\begin{table}[htbp]
    \centering
    \caption{Camera Pose Estimation results on TUM-dynamics, Sintel, and ScanNet datasets.}
    \label{tab:pose_evaluation}
    \resizebox{\textwidth}{!}{
    \begin{tabular}{l ccc ccc ccc}
        \toprule
        \multirow{2}{*}{\textbf{Method}} & \multicolumn{3}{c}{\textbf{{Sintel}}} & \multicolumn{3}{c}{\textbf{{TUM-dynamics}}} & \multicolumn{3}{c}{\textbf{ScanNet}} \\
        \cmidrule(lr){2-4} \cmidrule(lr){5-7} \cmidrule(lr){8-10}
         & \textbf{ATE $\downarrow$} & \textbf{RPE trans $\downarrow$} & \textbf{RPE rot $\downarrow$} & \textbf{ATE $\downarrow$} & \textbf{RPE trans $\downarrow$} & \textbf{RPE rot $\downarrow$} & \textbf{ATE $\downarrow$} & \textbf{RPE trans $\downarrow$} & \textbf{RPE rot $\downarrow$} \\
        \midrule
        CUT3R~\cite{wang2025continuous} & 0.210 & \underline{0.071} & 0.607 & 0.045 & 0.014 & 0.439 & 0.105 & 0.022 & 0.627 \\
        TTT3R~\cite{chen2025ttt3r} & 0.209 & 0.082 & 0.683 & 0.029 & 0.013 & 0.380 & 0.072 & 0.021 & 0.805 \\
        StreamVGGT~\cite{streamVGGT} & 0.232 & 0.082 & 0.669 & \underline{0.025} & \underline{0.012} & \underline{0.311} & 0.048 & 0.020 & 0.629 \\
        LongStream~\cite{cheng2026longstream} & \underline{0.194} & 0.076 & \underline{0.604} & 0.036 & 0.014 & 0.366 & \underline{0.037} & \underline{0.014} & \underline{0.431} \\
        \rowcolor{gray!10}
        \textbf{Ours} & \textbf{0.116} & \textbf{0.054} & \textbf{0.425} & \textbf{0.017} & \textbf{0.009} & \textbf{0.304} & \textbf{0.034} & \textbf{0.013} & \textbf{0.374} \\
        \bottomrule
    \end{tabular}
    }
\end{table}

\subsection{Ablation Study} 

To evaluate the effectiveness of our proposed components, we conduct ablation studies  within the Sparse Setting on 3D reconstruction on the 7-Scenes and NRGBD datasets. For fair comparison, we re-finetune VGGT~\cite{wang2025vggt} using the identical training datasets and settings as our method, denoted as StreamVGGT*.
Starting from this baseline (which employs the VGGT backbone and absolute pose supervision), we progressively incorporate our proposed designs: the frame-equivariant backbone ( $ \pi^3 $~\cite{wang2025pi}), the prefix-relative pose supervision, and the lookahead mechanism. The detailed results are summarized in ~\cref{tab:ablation_recon}, which highlight the following key findings:

\begin{table}[htbp]
    \centering
    \caption{Ablation study on 3D reconstruction under the Sparse Setting on NRGBD and 7-Scenes.}
    \label{tab:ablation_recon}
    \resizebox{\textwidth}{!}{
    \begin{tabular}{l l l c c c c c c c c c}
        \toprule
        \multicolumn{4}{c}{\textbf{Settings}} & \multicolumn{4}{c}{\textbf{NRGBD}} & \multicolumn{4}{c}{\textbf{7-Scenes}} \\
        \cmidrule(lr){1-4} \cmidrule(lr){5-8} \cmidrule(lr){9-12}
        \multirow{2}{*}{\textbf{Config}} & \multirow{2}{*}{\textbf{Backbone}} & \multirow{2}{*}{\textbf{Pose Loss}} & \multirow{2}{*}{\textbf{Lookahead}} & \multicolumn{2}{c}{\textbf{Acc}$\downarrow$} & \multicolumn{2}{c}{\textbf{Comp}$\downarrow$} & \multicolumn{2}{c}{\textbf{Acc}$\downarrow$} & \multicolumn{2}{c}{\textbf{Comp}$\downarrow$} \\
        \cmidrule(lr){5-6} \cmidrule(lr){7-8} \cmidrule(lr){9-10} \cmidrule(lr){11-12}
         & & & & \textbf{Mean} & \textbf{Med.} & \textbf{Mean} & \textbf{Med.} & \textbf{Mean} & \textbf{Med.} & \textbf{Mean} & \textbf{Med.} \\
        \midrule
        StreamVGGT~\cite{streamVGGT} & VGGT & Absolute & - 
            & 0.085 & 0.044 & 0.079 & 0.038
            & 0.132 & 0.058 & 0.116 & 0.042 \\ \hline
        StreamVGGT* & VGGT & Absolute & - 
            & 0.070 & 0.043 & 0.070 & 0.041
            & 0.101 & 0.063 & 0.106 & 0.067 \\
        w/o Rel.\& Look. & $\pi^3$ & Absolute & - 
            & 0.063 & 0.043 & 0.071 & 0.042
            & 0.064 & 0.040 & 0.083 & 0.059 \\
        w/o Look. & $\pi^3$ & Prefix-Relative & - 
            & 0.055 & 0.033 & 0.071 & 0.025
            & 0.054 & 0.032 & 0.069 & 0.048 \\
        \rowcolor{gray!10} 
        Ours (Full) & $\pi^3$ & Prefix-Relative & \checkmark 
            & 0.035 & 0.020 & 0.041 & 0.023
            & 0.052 & 0.032 & 0.069 & 0.048 \\
        \bottomrule
    \end{tabular}
    }
\end{table}

\textbf{Frame-equivariant backbone.}
Comparing Rows 2 and 3, we keep the supervision unchanged and only replace the VGGT~\cite{wang2025vggt} backbone and pretrained initialization, which uses a dedicated first-frame reference, with the frame-equivariant $\pi^3$~\cite{wang2025pi} backbone. This alone leads to a substantial improvement across both datasets. The result indicates that a frame-equivariant pretrained model is better matched to the causal streaming setting, where early frames typically contain the least geometric information. Unlike a first-frame-anchored design, it does not impose a privileged dependence on the earliest frame, thereby substantially mitigating the initial-frame bias discussed in \cref{sec.intro}.
  
\textbf{Prefix-Relative Supervision}: 
As shown in Row 4, switching from absolute to prefix-relative pose loss further improves reconstruction accuracy. This demonstrates that shifting the focus to relative geometric consistency imposes tighter constraints on the model, making learning less sensitive to deviations in early frames.

\textbf{Lookahead Mechanism}: 
The full model (Row 5) achieves the best performance by integrating the lookahead mechanism. The results confirm that the Lookahead module explicitly alleviates the challenge of information asymmetry in the streaming model by leveraging future context to refine history estimates.

\section{Conclusion}

In this paper, we present Stable3R, a holistic framework designed to address the inherent instability in streaming 3D reconstruction caused by early-frame bias. By identifying the critical bottleneck of over-reliance on information-poor initial estimates, we develop two complementary designs: a lookahead attention mechanism to enrich historical caches with future context, and a frame-equivariant architecture with prefix-relative pose supervision to stabilize geometric references.
Comprehensive experiments across various benchmarks demonstrate that our method achieves superior reconstruction quality and camera pose estimation accuracy. Notably, Stable3R exhibits robust generalization, maintaining high performance even in challenging settings such as sparse views and long-sequence inference.

\begin{ack}
This work was conducted during Jincheng Xiong's internship
at AMAP, Alibaba. We thank Linzhe Shi and Xin Zhang for their valuable discussions
and feedback.
\end{ack}

{
    \small
    \bibliographystyle{plainnat}

    \bibliography{reference}
}

\newpage
\appendix

\section{Additional Experiments and Analysis}
\label{sec:appendix_exp}

In this section, we provide additional experiments and analyses to further
investigate the stability of geometric references, the behavior of lookahead
attention, and the efficiency and scalability of Stable3R. We first extend the
early-frame bias analysis in the main paper to different datasets and
perturbation settings. We then analyze the lookahead mechanism from the
perspectives of update range, geometric interpretation, and long-horizon
stability. Finally, we report detailed runtime and memory measurements,
additional long-sequence experiments, and comparisons with concurrent methods.

\subsection{Additional Analysis of Early-Frame Bias}
\label{sec:app_early_bias}

\subsubsection{Stability across Prefix Lengths and Datasets}
\label{sec:app_prefix_stability}

We first extend the geometric context probe introduced in
\cref{fig:teaser} to NRGBD, 7-Scenes, TUM-dynamics, and Sintel.
Following the same protocol, we fix the initial and probe frames while varying
the number of intermediate observations from 0 to 8. We report the mean
Absolute Trajectory Error (ATE) of the probe frames together with the ATE range
across different prefix configurations.

As shown in \cref{tab:probe_prefix}, StreamVGGT exhibits noticeable variation
as the number of intermediate observations changes, particularly on NRGBD,
7-Scenes, and Sintel. In contrast, Stable3R consistently achieves lower
probe-frame ATE and substantially smaller variation across prefix lengths.
These results indicate that the early-frame bias observed in
\cref{fig:teaser} generalizes across datasets, and that Stable3R provides more
stable geometric references under varying amounts of historical context.

\begin{table}[!hbpt]
    \centering
    \caption{
    Geometric context probe across different datasets and prefix lengths.
    We fix the initial and probe frames while varying the number of intermediate
    frames. Lower ATE and a smaller ATE range indicate more stable historical
    geometric references.
    }
    \label{tab:probe_prefix}
    \setlength{\tabcolsep}{4.2pt}
    \resizebox{\textwidth}{!}
    {
    \begin{tabular}{llcccccc}
        \toprule
        \textbf{Dataset} & \textbf{Method}
        & \textbf{0 frame}
        & \textbf{2 frames}
        & \textbf{4 frames}
        & \textbf{6 frames}
        & \textbf{8 frames}
        & \textbf{ATE Range}$\downarrow$ \\
        \midrule
        \multirow{2}{*}{NRGBD}
        & StreamVGGT & 0.13539 & 0.12477 & 0.11450 & 0.10707 & 0.10022 & 0.03517 \\
        & Stable3R   & \textbf{0.04137} & \textbf{0.03182} & \textbf{0.03105}
                     & \textbf{0.02992} & \textbf{0.03064} & \textbf{0.01145} \\
        \midrule

        \multirow{2}{*}{TUM-dynamics}
        & StreamVGGT & 0.01676 & 0.01556 & 0.01513 & 0.01507 & 0.01496 & 0.00180 \\
        & Stable3R   & \textbf{0.01378} & \textbf{0.01300} & \textbf{0.01317}
                     & \textbf{0.01300} & \textbf{0.01325} & \textbf{0.00078} \\
        \midrule

        \multirow{2}{*}{7-Scenes}
        & StreamVGGT & 0.12108 & 0.12250 & 0.09757 & 0.08873 & 0.07615 & 0.04635 \\
        & Stable3R   & \textbf{0.07418} & \textbf{0.07319} & \textbf{0.07335}
                     & \textbf{0.05919} & \textbf{0.05821} & \textbf{0.01597} \\
        \midrule

        \multirow{2}{*}{Sintel}
        & StreamVGGT & 0.14975 & 0.17687 & 0.14751 & 0.13530 & 0.13171 & 0.04516 \\
        & Stable3R   & \textbf{0.06194} & \textbf{0.06104} & \textbf{0.06255}
                     & \textbf{0.06512} & \textbf{0.06416} & \textbf{0.00408} \\
        \bottomrule
    \end{tabular}}
\end{table}

\subsubsection{Sensitivity to Initial-Frame Quality}
\label{sec:app_initial_quality}

We next directly evaluate the sensitivity to the quality of the initial
observation. Specifically, we apply Gaussian blur only to the first frame while
keeping all subsequent observations unchanged, and measure the resulting
increase in probe-frame ATE. This perturbation isolates the influence of the
initial observation without modifying the remaining sequence.

As shown in \cref{tab:first_frame_quality}, degrading the first frame
consistently leads to a larger ATE increase for StreamVGGT across all four
datasets. Stable3R is substantially less affected by the same perturbation,
indicating reduced dependence on the quality of the initial estimate. This
provides direct evidence that alleviating first-frame anchoring improves
streaming stability.

\begin{table}[!hbpt]
    \centering
    \caption{
    Sensitivity to the quality of the initial frame. We apply Gaussian blur
    only to the first frame and report the increase in probe-frame ATE.
    Lower values indicate weaker dependence on the initial observation.
    }
    \label{tab:first_frame_quality}
    \setlength{\tabcolsep}{5pt}
    \begin{tabular}{llcccc}
        \toprule
        \textbf{Dataset} & \textbf{Method}
        & \textbf{2 frames}
        & \textbf{4 frames}
        & \textbf{6 frames}
        & \textbf{8 frames} \\
        \midrule
        \multirow{2}{*}{NRGBD}
        & StreamVGGT & 0.01642 & 0.01138 & 0.00741 & 0.00708 \\
        & Stable3R   & \textbf{0.00006} & \textbf{0.00160}
                     & \textbf{0.00237} & \textbf{0.00198} \\
        \midrule

        \multirow{2}{*}{TUM-dynamics}
        & StreamVGGT & 0.00651 & 0.00495 & 0.00363 & 0.00235 \\
        & Stable3R   & \textbf{0.00081} & \textbf{0.00037}
                     & \textbf{0.00029} & \textbf{0.00056} \\
        \midrule

        \multirow{2}{*}{7-Scenes}
        & StreamVGGT & 0.00531 & 0.00599 & 0.00861 & 0.00694 \\
        & Stable3R   & \textbf{0.00040} & \textbf{0.00239}
                     & \textbf{0.00098} & \textbf{0.00037} \\
        \midrule

        \multirow{2}{*}{Sintel}
        & StreamVGGT & 0.01992 & 0.03024 & 0.03649 & 0.03766 \\
        & Stable3R   & \textbf{0.01097} & \textbf{0.00095}
                     & \textbf{0.00202} & \textbf{0.00292} \\
        \bottomrule
    \end{tabular}
\end{table}

\subsubsection{Sensitivity to Historical Frame Overlap}
\label{sec:app_overlap}

Besides prefix length and initial-frame quality, we further study whether
geometric stability depends on the overlap of historical observations. We fix
the initial and probe frames while varying the intermediate frames. Overlap is
quantified according to the visibility of scene points in the probe view, and
the samples are divided into three equally sized groups corresponding to low,
medium, and high overlap.

As reported in \cref{tab:overlap_stability}, Stable3R achieves lower
probe-frame ATE across all overlap levels and generally exhibits smaller
variation among the three groups. This indicates that Stable3R is less
sensitive to the composition of intermediate observations and provides more
stable historical geometric references.

\begin{table}[!htbp]
    \centering
    \caption{
    Sensitivity to the overlap of intermediate observations. Lower ATE and a
    smaller ATE range indicate greater stability under changes in historical
    frame composition.
    }
    \label{tab:overlap_stability}
    \setlength{\tabcolsep}{5pt}
    \begin{tabular}{llcccc}
        \toprule
        \textbf{Dataset} & \textbf{Method}
        & \textbf{Low}
        & \textbf{Medium}
        & \textbf{High}
        & \textbf{ATE Range}$\downarrow$ \\
        \midrule
        \multirow{2}{*}{NRGBD}
        & StreamVGGT & 0.1341 & 0.1346 & 0.1315 & 0.0031 \\
        & Stable3R   & \textbf{0.0363} & \textbf{0.0367}
                     & \textbf{0.0363} & \textbf{0.0004} \\
        \midrule

        \multirow{2}{*}{TUM-dynamics}
        & StreamVGGT & 0.0162 & 0.0168 & 0.0168 & 0.0006 \\
        & Stable3R   & \textbf{0.0138} & \textbf{0.0136}
                     & \textbf{0.0135} & \textbf{0.0003} \\
        \midrule

        \multirow{2}{*}{7-Scenes}
        & StreamVGGT & 0.1167 & 0.1030 & 0.1009 & 0.0158 \\
        & Stable3R   & \textbf{0.0710} & \textbf{0.0657}
                     & \textbf{0.0570} & \textbf{0.0140} \\
        \midrule

        \multirow{2}{*}{Sintel}
        & StreamVGGT & 0.1414 & 0.1381 & 0.1461 & 0.0080 \\
        & Stable3R   & \textbf{0.0634} & \textbf{0.0666}
                     & \textbf{0.0626} & \textbf{0.0040} \\
        \bottomrule
    \end{tabular}
\end{table}

\subsection{Analysis of Lookahead Attention}
\label{sec:app_lookahead_analysis}

We next analyze the behavior of lookahead attention in greater detail. In
particular, we study how much historical context needs to be updated, whether
the learned lookahead representation correlates with geometric properties, and
whether its recursive accumulation remains stable over long sequences.

\subsubsection{Sensitivity to Lookahead Update Range}
\label{sec:app_lookahead_range}

We first study how many historical frames need to be refined by lookahead
attention. In addition to the default setting that updates the full history, we
restrict the lookahead update to only the 10 most recent historical frames, or
disable historical updates entirely.

As shown in \cref{tab:lookahead_range}, updating only the most recent 10 frames
results in nearly identical reconstruction quality to full-history refinement
on both NRGBD and 7-Scenes. In contrast, disabling historical refinement leads
to a clear performance drop, particularly on NRGBD. This suggests that most of
the benefit of lookahead attention is captured by refining relatively recent
historical representations, while increasingly distant frames provide only
limited additional gains.

\begin{table}[!htbp]
    \centering
    \caption{
    Sensitivity to the number of historical frames updated by lookahead
    attention under the Mid-length setting.
    }
    \label{tab:lookahead_range}
    \setlength{\tabcolsep}{3.8pt}
    \resizebox{\textwidth}{!}{
    \begin{tabular}{llcccccc}
        \toprule
        \textbf{Dataset} & \textbf{Method}
        & \textbf{Acc Mean}$\downarrow$
        & \textbf{Acc Med.}$\downarrow$
        & \textbf{Comp Mean}$\downarrow$
        & \textbf{Comp Med.}$\downarrow$
        & \textbf{NC Mean}$\uparrow$
        & \textbf{NC Med.}$\uparrow$ \\
        \midrule

        \multirow{3}{*}{7-Scenes}
        & Stable3R                 & {0.019} & {0.010}
                                    & {0.022} & {0.010}
                                    & {0.722} & {0.834} \\
        & Stable3R (update 10)     & {0.019} & {0.010}
                                    & {0.022} & {0.010}
                                    & 0.721 & {0.834} \\
        & Stable3R (update 0)      & 0.023 & 0.012 & 0.023 & 0.011 & 0.721 & {0.835} \\
        \midrule

        \multirow{3}{*}{NRGBD}
        & Stable3R                 & {0.022} & {0.014}
                                    & {0.020} & {0.010}
                                    & {0.884} & {0.987} \\
        & Stable3R (update 10)     & 0.023 & {0.014}
                                    & 0.021 & {0.010}
                                    & {0.884} & 0.986 \\
        & Stable3R (update 0)      & 0.034 & 0.022 & 0.029 & 0.013 & 0.863 & 0.983 \\
        \bottomrule
    \end{tabular}}
\end{table}

\subsubsection{Geometric Interpretation of Lookahead Attention}
\label{sec:app_lookahead_geometry}

We further investigate whether the learned lookahead representation reflects
meaningful geometric properties of the historical observations. We analyze it
from two complementary perspectives: frame-level co-visibility and token-level
geometric reliability.

\paragraph{Co-visibility.}
For each current--historical frame pair, we estimate co-visibility through
point-cloud projection and divide all pairs into equally sized low-, medium-,
and high-co-visibility groups. We then measure the proportion of historical
attention weights that increase after applying the lookahead term.

As shown in \cref{tab:lookahead_covisibility}, highly co-visible historical
frames are more likely to receive increased relative attention than
low-overlap frames. This trend is consistent on both 7-Scenes and NRGBD,
suggesting that lookahead preferentially preserves geometrically relevant
historical observations.

\begin{table}[!htbp]
    \centering
    \caption{
    Relationship between frame co-visibility and lookahead attention.
    We report the proportion of historical attention weights that increase
    after applying the lookahead term.
    }
    \label{tab:lookahead_covisibility}
    \setlength{\tabcolsep}{9pt}
    \begin{tabular}{lccc}
        \toprule
        \textbf{Dataset}
        & \textbf{Low co-vis.}
        & \textbf{Medium co-vis.}
        & \textbf{High co-vis.} \\
        \midrule
        7-Scenes & 31.1\% & 33.4\% & {39.5\%} \\
        NRGBD    & 26.4\% & 32.3\% & {39.2\%} \\
        \bottomrule
    \end{tabular}
\end{table}

\paragraph{Geometric reliability.}
For each historical token, we compute its predicted point-cloud error and mean
lookahead bias. We divide all tokens into five equally sized groups according
to their geometric error and report the average bias relative to the
lowest-error group.

As shown in \cref{tab:lookahead_error_bias}, the lookahead bias increases
consistently with geometric error. Since the lookahead term is subtracted from
the causal attention logits in \cref{equ:lookahead_attenion}, historical tokens with
larger geometric errors receive stronger suppression. Together with the
co-visibility analysis above, these results show that the learned lookahead
representation correlates with both geometric relevance and reliability.

\begin{table}[!htbp]
    \centering
    \caption{
    Relationship between historical-token geometric error and lookahead bias.
    Tokens are equally divided into five groups according to point-cloud error.
    The relative bias is normalized by the mean bias of the lowest-error group.
    }
    \label{tab:lookahead_error_bias}
    \setlength{\tabcolsep}{5pt}
    \resizebox{0.82\textwidth}{!}{
    \begin{tabular}{llccccc}
        \toprule
        \textbf{Dataset} & \textbf{Metric}
        & \textbf{Group 1}
        & \textbf{Group 2}
        & \textbf{Group 3}
        & \textbf{Group 4}
        & \textbf{Group 5} \\
        \midrule

        \multirow{2}{*}{7-Scenes}
        & Point-cloud error
        & 3.92 & 5.72 & 7.27 & 9.13 & 16.87 \\
        & Relative bias
        & 100.0\% & 104.5\% & 107.9\% & 113.2\% & 123.5\% \\
        \midrule

        \multirow{2}{*}{NRGBD}
        & Point-cloud error
        & 2.80 & 4.09 & 5.15 & 6.72 & 15.43 \\
        & Relative bias
        & 100.0\% & 105.8\% & 109.8\% & 120.3\% & 133.3\% \\
        \bottomrule
    \end{tabular}}
\end{table}

\subsubsection{Long-Horizon Stability of Lookahead States}
\label{sec:app_lookahead_stability}

The recursive formulation in \cref{equ:update_equation} continuously
accumulates newly observed information into the lookahead state. We therefore
analyze whether this accumulation leads to an increasingly dominant or
unstable attention bias over long sequences.

We track the RMS norm of the first-frame lookahead state $U$ and the attention
assigned to the first frame as the sequence length increases from 4 to 128
frames. As shown in \cref{tab:lookahead_stability}, the magnitude of the
lookahead state grows approximately linearly with sequence length. However,
its effect on normalized attention remains stable: lookahead consistently
reduces the attention assigned to the first frame by approximately
20--24\% across all evaluated lengths.

These results indicate that although the internal lookahead representation
accumulates over time, its influence on normalized attention does not become
increasingly dominant within the evaluated sequence range.

\begin{table}[!htbp]
    \centering
    \caption{
    Long-horizon behavior of the first-frame lookahead state. We report its RMS
    norm together with the attention assigned to the first frame with and
    without lookahead.
    }
    \label{tab:lookahead_stability}
    \setlength{\tabcolsep}{5pt}
    \begin{tabular}{ccccc}
        \toprule
        \textbf{Seq. Length}
        & \textbf{$U$ RMS}
        & \textbf{Attn. w/ Lookahead}
        & \textbf{Attn. w/o Lookahead}
        & \textbf{Relative Change} \\
        \midrule
        4   & 183   & 19.371\% & 25.239\% & -23.2\% \\
        8   & 424   & 9.734\%  & 12.744\% & -23.6\% \\
        16  & 900   & 4.804\%  & 6.238\%  & -23.0\% \\
        32  & 1,849 & 2.406\%  & 3.102\%  & -22.4\% \\
        64  & 3,806 & 1.271\%  & 1.608\%  & -21.0\% \\
        128 & 7,608 & 0.666\%  & 0.835\%  & -20.2\% \\
        \bottomrule
    \end{tabular}
\end{table}

\subsubsection{Replaying Earlier Frames with Refined Lookahead States}
\label{sec:app_replay}

Stable3R refines historical cache representations as new observations arrive,
but does not revise previously emitted predictions. To further examine this
design, we perform a replay experiment in which the lookahead states obtained
after a complete online pass are retained and used to reprocess earlier
frames.

As shown in \cref{tab:lookahead_replay}, replaying earlier frames with the
enriched states does not improve reconstruction and instead degrades
performance. We attribute this to a mismatch between enriched historical
states and early-frame queries that is not encountered during training. This
result further clarifies that Stable3R is designed to improve subsequent
online predictions through refined historical representations, rather than to
retroactively revise previous outputs.

\begin{table}[!htbp]
    \centering
    \caption{
    Reconstruction results when replaying the sequence with lookahead states
    enriched during the first online pass.
    }
    \label{tab:lookahead_replay}
    \setlength{\tabcolsep}{4.5pt}
    \resizebox{0.9\textwidth}{!}{
    \begin{tabular}{llcccccc}
        \toprule
        \textbf{Dataset} & \textbf{Setting}
        & \textbf{Acc}$\downarrow$
        & \textbf{Acc Med.}$\downarrow$
        & \textbf{Comp}$\downarrow$
        & \textbf{Comp Med.}$\downarrow$
        & \textbf{NC}$\uparrow$
        & \textbf{NC Med.}$\uparrow$ \\
        \midrule

        \multirow{2}{*}{7-Scenes}
        & Stable3R          & {0.019} & {0.010}
                            & 0.022 & 0.010 & {0.722} & {0.834} \\
        & Enriched-$U$ replay
                            & 0.022 & 0.012
                            & {0.022} & {0.009}
                            & 0.721 & {0.834} \\
        \midrule

        \multirow{2}{*}{NRGBD}
        & Stable3R          & {0.022} & {0.014}
                            & {0.020} & {0.010}
                            & {0.884} & {0.987} \\
        & Enriched-$U$ replay
                            & 0.032 & 0.023 & 0.026 & 0.014 & 0.860 & 0.981 \\
        \bottomrule
    \end{tabular}}
\end{table}

\subsection{Streaming Efficiency and Scalability}
\label{sec:app_efficiency}

Since lookahead attention augments the conventional KV cache with additional
states, we provide a detailed analysis of its computational and memory
overhead. All efficiency experiments in this section are measured on a single
NVIDIA A100-SXM4-80GB GPU.

\subsubsection{Runtime and Memory Overhead of Lookahead Attention}
\label{sec:app_lookahead_efficiency}

The QU-KV cache stores two additional lookahead states compared with standard
KV-cached causal attention, introducing approximately 50 MiB of additional
memory per frame. The sequential execution of lookahead update and attention
also incurs additional latency.

We compare Stable3R with Stable3R without lookahead and StreamVGGT in
\cref{tab:lookahead_efficiency}. As expected, removing lookahead reduces both
runtime and memory consumption. Nevertheless, the full Stable3R model remains
more efficient than StreamVGGT across all evaluated sequence lengths. These
results characterize the accuracy--efficiency trade-off introduced by
lookahead refinement.

\begin{table}[!htbp]
    \centering
    \caption{
    Runtime and memory comparison for unbounded-memory streaming inference.
    All results are measured on a single A100-SXM4-80GB GPU.
    }
    \label{tab:lookahead_efficiency}
    \setlength{\tabcolsep}{4pt}
    \resizebox{\textwidth}{!}{
    \begin{tabular}{llcccc}
        \toprule
        \textbf{Method} & \textbf{Metric}
        & \textbf{50 frames}
        & \textbf{150 frames}
        & \textbf{200 frames}
        & \textbf{300 frames} \\
        \midrule

        \multirow{3}{*}{Stable3R}
        & FPS$\uparrow$          & 7.98 & 3.84 & 3.08 & 2.17 \\
        & Latency (ms)$\downarrow$ & 189.35 & 463.78 & 594.34 & 859.40 \\
        & Mem. (GiB)$\downarrow$ & 9.54 & 19.49 & 24.53 & 34.43 \\
        \midrule

        \multirow{3}{*}{Stable3R w/o Lookahead}
        & FPS$\uparrow$          & {11.75} & {7.01} & {5.93} & {3.85} \\
        & Latency (ms)$\downarrow$ & {104.70} & {234.23}
                                      & {302.46} & {432.31} \\
        & Mem. (GiB)$\downarrow$ & {7.05} & {12.07}
                                   & {15.04} & {20.78} \\
        \midrule

        \multirow{3}{*}{StreamVGGT}
        & FPS$\uparrow$          & 5.10 & 2.56 & 2.28 & 2.01 \\
        & Latency (ms)$\downarrow$ & 259.93 & 604.13 & 780.55 & 1133.37 \\
        & Mem. (GiB)$\downarrow$ & 13.08 & 25.65 & 31.93 & 40.74 \\
        \bottomrule
    \end{tabular}}
\end{table}

\subsubsection{Efficiency of the Triton Implementation}
\label{sec:app_triton}

We further evaluate the practical benefit of the customized Triton
implementation introduced in \cref{sec:lookahead_attention}. We compare it with a
straightforward PyTorch implementation of the same lookahead computation and
report forward latency and peak GPU memory.

As shown in \cref{tab:triton_efficiency}, the Triton implementation
substantially reduces both runtime and memory consumption, with the benefit
becoming increasingly pronounced for longer sequences. At 64 frames, the
forward latency decreases from 1058.8\,ms to 342.0\,ms, while peak memory
decreases from 9.13\,GiB to 1.06\,GiB. These results validate the effectiveness
of the block-wise parallelization and memory-efficient tiling strategy.

\begin{table}[!htbp]
    \centering
    \caption{
    Efficiency comparison between the PyTorch and Triton implementations of
    lookahead attention.
    }
    \label{tab:triton_efficiency}
    \setlength{\tabcolsep}{5pt}
    \begin{tabular}{llccccc}
        \toprule
        \textbf{Metric} & \textbf{Impl.}
        & \textbf{4}
        & \textbf{8}
        & \textbf{16}
        & \textbf{32}
        & \textbf{64} \\
        \midrule

        \multirow{2}{*}{Forward time (ms)$\downarrow$}
        & PyTorch & 5.66 & 18.58 & 67.74 & 262.25 & 1058.80 \\
        & Triton  & \textbf{1.96} & \textbf{4.60} & \textbf{17.16}
                  & \textbf{73.00} & \textbf{342.00} \\
        \midrule

        \multirow{2}{*}{Peak memory (GiB)$\downarrow$}
        & PyTorch & 0.53 & 1.10 & 2.25 & 4.54 & 9.13 \\
        & Triton  & \textbf{0.07} & \textbf{0.13} & \textbf{0.27}
                  & \textbf{0.53} & \textbf{1.06} \\
        \bottomrule
    \end{tabular}
\end{table}

\subsubsection{Bounded-Memory Long-Sequence Inference}
\label{sec:app_bounded_efficiency}

For practical long-sequence deployment, we additionally combine Stable3R with
the frame-level bounded-memory retention strategy used in
\cite{xu2026framevggt}. We use a fixed 20-frame memory budget consisting of
16 recent frames and 4 sparse anchors. For brevity, we denote this
frame-level retention mechanism as FERM in the following tables.

As shown in \cref{tab:direct_ferm_scaling}, direct inference exhibits
approximately linear growth in cache size, GPU memory, and latency as the
sequence becomes longer. In contrast, FERM keeps the cache size fixed at
approximately 2.61\,GiB. Across sequences from 25 to 300 frames, peak GPU
memory remains below 8.68\,GiB and latency remains between 109.7 and
140.1\,ms.

\begin{table}[!htbp]
    \centering
    \caption{
    Scaling behavior of direct and bounded-memory inference as the sequence
    length increases.
    }
    \label{tab:direct_ferm_scaling}
    \setlength{\tabcolsep}{2.7pt}
    {
    \begin{tabular}{c|cccc|cccc}
        \toprule
        \multirow{2}{*}{\textbf{Frames}}
        & \multicolumn{4}{c|}{\textbf{Direct}}
        & \multicolumn{4}{c}{\textbf{FERM}} \\
        \cmidrule(lr){2-5}\cmidrule(lr){6-9}
        & \textbf{Latency}
        & \textbf{FPS}
        & \textbf{Mem.}
        & \textbf{Cache}
        & \textbf{Latency}
        & \textbf{FPS}
        & \textbf{Mem.}
        & \textbf{Cache} \\
        & (ms)$\downarrow$
        & $\uparrow$
        & (GiB)$\downarrow$
        & (GiB)$\downarrow$
        & (ms)$\downarrow$
        & $\uparrow$
        & (GiB)$\downarrow$
        & (GiB)$\downarrow$ \\
        \midrule
        25  & 117.36 & 10.891 & 7.09  & 2.49
            & 109.7  & 9.09   & 8.01  & 2.61 \\
        50  & 189.35 & 7.981  & 9.54  & 4.98
            & 111.4  & 8.98   & 8.25  & 2.61 \\
        100 & 322.31 & 5.228  & 14.54 & 9.96
            & 114.8  & 8.77   & 8.33  & 2.61 \\
        150 & 463.78 & 3.839  & 19.49 & 14.93
            & 132.1  & 7.57   & 8.34  & 2.61 \\
        200 & 594.34 & 3.075  & 24.53 & 19.91
            & 136.6  & 7.32   & 8.43  & 2.61 \\
        250 & 733.64 & 2.531  & 29.45 & 24.89
            & 140.0  & 7.11   & 8.51  & 2.61 \\
        300 & 859.40 & 2.169  & 34.43 & 29.87
            & 140.1  & 7.14   & 8.68  & 2.61 \\
        \bottomrule
    \end{tabular}}
\end{table}

We further compare the bounded-memory efficiency of Stable3R against OVGGT,
FrameVGGT, and InfiniteVGGT. As shown in
\cref{tab:bounded_baseline_efficiency}, Stable3R achieves throughput comparable
to OVGGT while exhibiting a flatter GPU-memory growth curve. FrameVGGT and
InfiniteVGGT require substantially more GPU memory and achieve lower
throughput under the evaluated setting.

\begin{table}[!hbpt]
    \centering
    \caption{
    Efficiency comparison with bounded-memory streaming baselines.
    }
    \label{tab:bounded_baseline_efficiency}
    \setlength{\tabcolsep}{4pt}
    {
    \begin{tabular}{llcccc}
        \toprule
        \textbf{Method} & \textbf{Metric}
        & \textbf{50 frames}
        & \textbf{150 frames}
        & \textbf{200 frames}
        & \textbf{300 frames} \\
        \midrule

        \multirow{3}{*}{Stable3R (w/ FERM)}
        & FPS$\uparrow$ & 8.98 & 7.57 & 7.32 & 7.14 \\
        & Latency (ms)$\downarrow$ & 111.4 & 132.1 & 136.6 & 140.1 \\
        & Mem. (GiB)$\downarrow$ & {8.25} & {8.34}
                                  & {8.43} & {8.68} \\
        \midrule

        \multirow{3}{*}{OVGGT}
        & FPS$\uparrow$ & 8.49 & {8.20} & {8.18} & {8.13} \\
        & Latency (ms)$\downarrow$ & 117.8 & {122.0} & {122.3} & {123.0} \\
        & Mem. (GiB)$\downarrow$ & 8.65 & 9.57 & 10.03 & 10.96 \\
        \midrule

        \multirow{3}{*}{FrameVGGT}
        & FPS$\uparrow$ & 6.14 & 5.81 & 5.94 & 5.80 \\
        & Latency (ms)$\downarrow$ & 162.9 & 172.1 & 168.4 & 172.4 \\
        & Mem. (GiB)$\downarrow$ & 15.07 & 17.39 & 18.24 & 19.17 \\
        \midrule

        \multirow{3}{*}{InfiniteVGGT}
        & FPS$\uparrow$ & 5.53 & 3.91 & 3.68 & 3.56 \\
        & Latency (ms)$\downarrow$ & 180.8 & 255.8 & 271.7 & 280.9 \\
        & Mem. (GiB)$\downarrow$ & 15.21 & 16.28 & 16.77 & 17.71 \\
        \bottomrule
    \end{tabular}}
\end{table}

\subsection{Additional Long-Sequence Evaluation}
\label{sec:app_long_sequence}

\subsubsection{Effect of the Frame-Level Retention Strategy}
\label{sec:app_retention_ablation}

The long-sequence evaluation in the main paper adopts FERM to control
inference cost. To distinguish the contribution of Stable3R itself from that
of the external memory-retention mechanism, we additionally evaluate Stable3R
by directly processing the same 300-frame sequences without bounded-memory
retention.

As shown in \cref{tab:retention_ablation}, direct inference achieves better
reconstruction accuracy on most metrics than the bounded-memory variant. This
indicates that the long-sequence reconstruction gains primarily originate from
Stable3R rather than from FERM. The retention mechanism mainly serves as an
efficient extension that substantially reduces memory and computation while
preserving comparable reconstruction quality.

\begin{table}[!htbp]
    \centering
    \caption{
    Effect of the frame-level retention strategy on 300-frame reconstruction.
    Direct inference processes the full history, while FERM uses bounded
    memory.
    }
    \label{tab:retention_ablation}
    \setlength{\tabcolsep}{2.9pt}
    \resizebox{\textwidth}{!}{
    \begin{tabular}{lcccccccccccc}
        \toprule
        \multirow{3}{*}{\textbf{Method}}
        & \multicolumn{6}{c}{\textbf{7-Scenes}}
        & \multicolumn{6}{c}{\textbf{NRGBD}} \\
        \cmidrule(lr){2-7}\cmidrule(lr){8-13}
        & \multicolumn{2}{c}{\textbf{Acc}$\downarrow$}
        & \multicolumn{2}{c}{\textbf{Comp}$\downarrow$}
        & \multicolumn{2}{c}{\textbf{NC}$\uparrow$}
        & \multicolumn{2}{c}{\textbf{Acc}$\downarrow$}
        & \multicolumn{2}{c}{\textbf{Comp}$\downarrow$}
        & \multicolumn{2}{c}{\textbf{NC}$\uparrow$} \\
        & Mean & Med. & Mean & Med. & Mean & Med.
        & Mean & Med. & Mean & Med. & Mean & Med. \\
        \midrule

        Stable3R (w/ FERM)
        & 0.020 & 0.010
        & \textbf{0.014} & \textbf{0.003}
        & \textbf{0.572} & 0.610
        & 0.029 & 0.018
        & 0.009 & 0.003
        & 0.627 & 0.707 \\

        Stable3R (w/o FERM)
        & \textbf{0.013} & \textbf{0.005}
        & 0.015 & 0.004
        & \textbf{0.572} & \textbf{0.611}
        & \textbf{0.018} & \textbf{0.011}
        & \textbf{0.007} & \textbf{0.002}
        & \textbf{0.645} & \textbf{0.738} \\
        \bottomrule
    \end{tabular}}
\end{table}

\subsubsection{Evaluation on Long3D}
\label{sec:app_long3d}

We additionally evaluate Stable3R on the Long3D benchmark using the same
bounded-memory inference strategy. As shown in \cref{tab:long3d}, Stable3R
improves reconstruction accuracy from 2.09 to 1.72 and normal consistency from
0.52 to 0.54 compared with InfiniteVGGT.

Stable3R obtains worse completeness under this evaluation. We observe that the
two methods use different effective fields of view: Stable3R crops image
borders during inference, which removes part of the reconstructed point cloud
and affects the completeness metric. Nevertheless, the results demonstrate
that Stable3R can be extended to substantially longer sequences without
numerical instability.

\begin{table}[!htbp]
    \centering
    \caption{Long-sequence reconstruction results on the Long3D benchmark.}
    \label{tab:long3d}
    \setlength{\tabcolsep}{10pt}
    \begin{tabular}{lccc}
        \toprule
        \textbf{Method}
        & \textbf{Acc}$\downarrow$
        & \textbf{Comp}$\downarrow$
        & \textbf{NC}$\uparrow$ \\
        \midrule
        InfiniteVGGT & 2.09 & \textbf{0.85} & 0.52 \\
        Stable3R (w/ FERM)     & \textbf{1.72} & 1.84 & \textbf{0.54} \\
        \bottomrule
    \end{tabular}
\end{table}

\section{Technical appendices and supplementary material}

\begin{figure}[h]
\centering
\includegraphics[width=\linewidth]{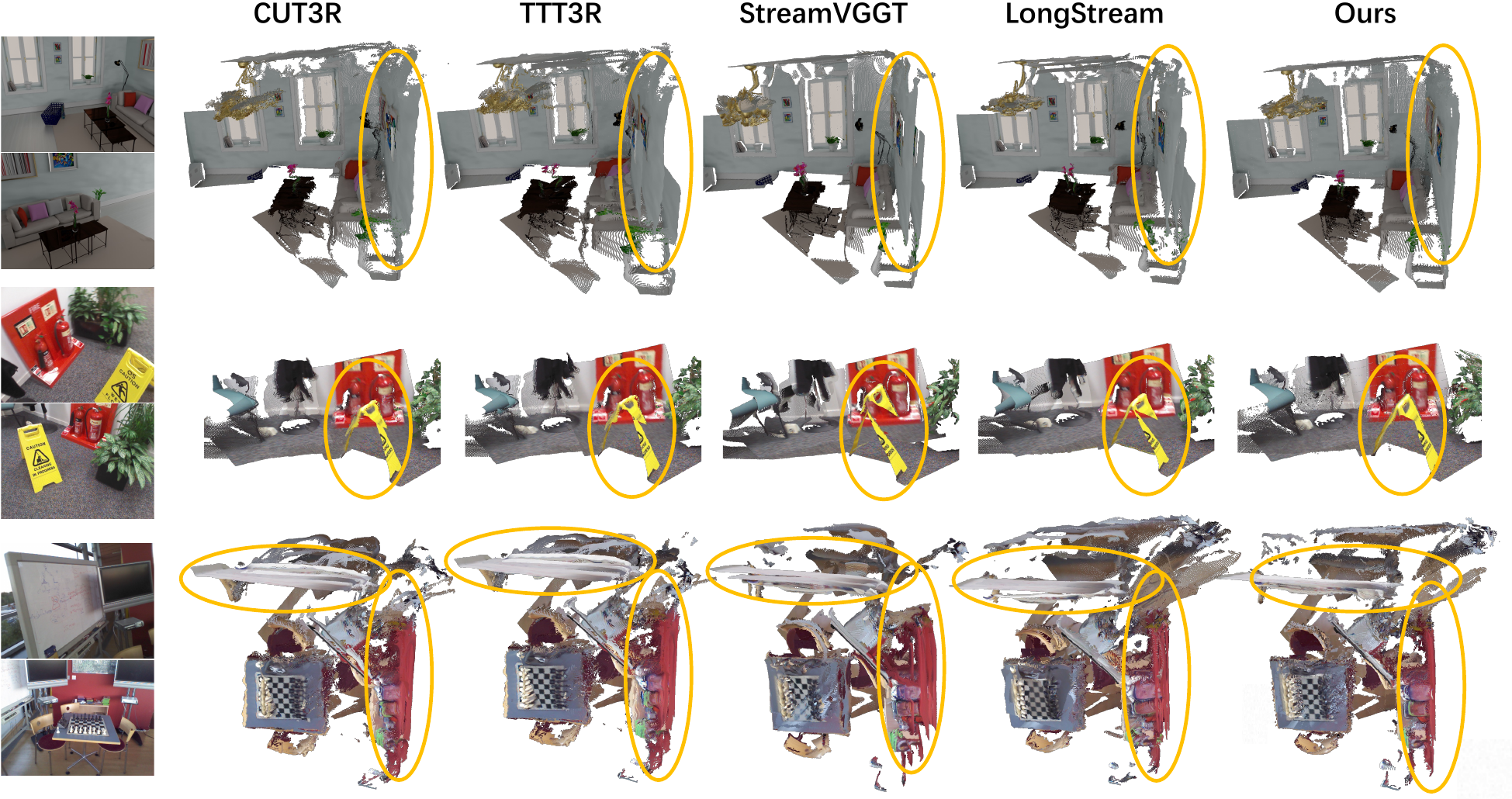}
\caption{
Qualitative Reconstruction Results on the NRGBD and 7-scenes datasets on mid-length settings.
}
\label{fig:app3}
\end{figure}

\begin{figure}[h]
\centering
\includegraphics[width=\linewidth]{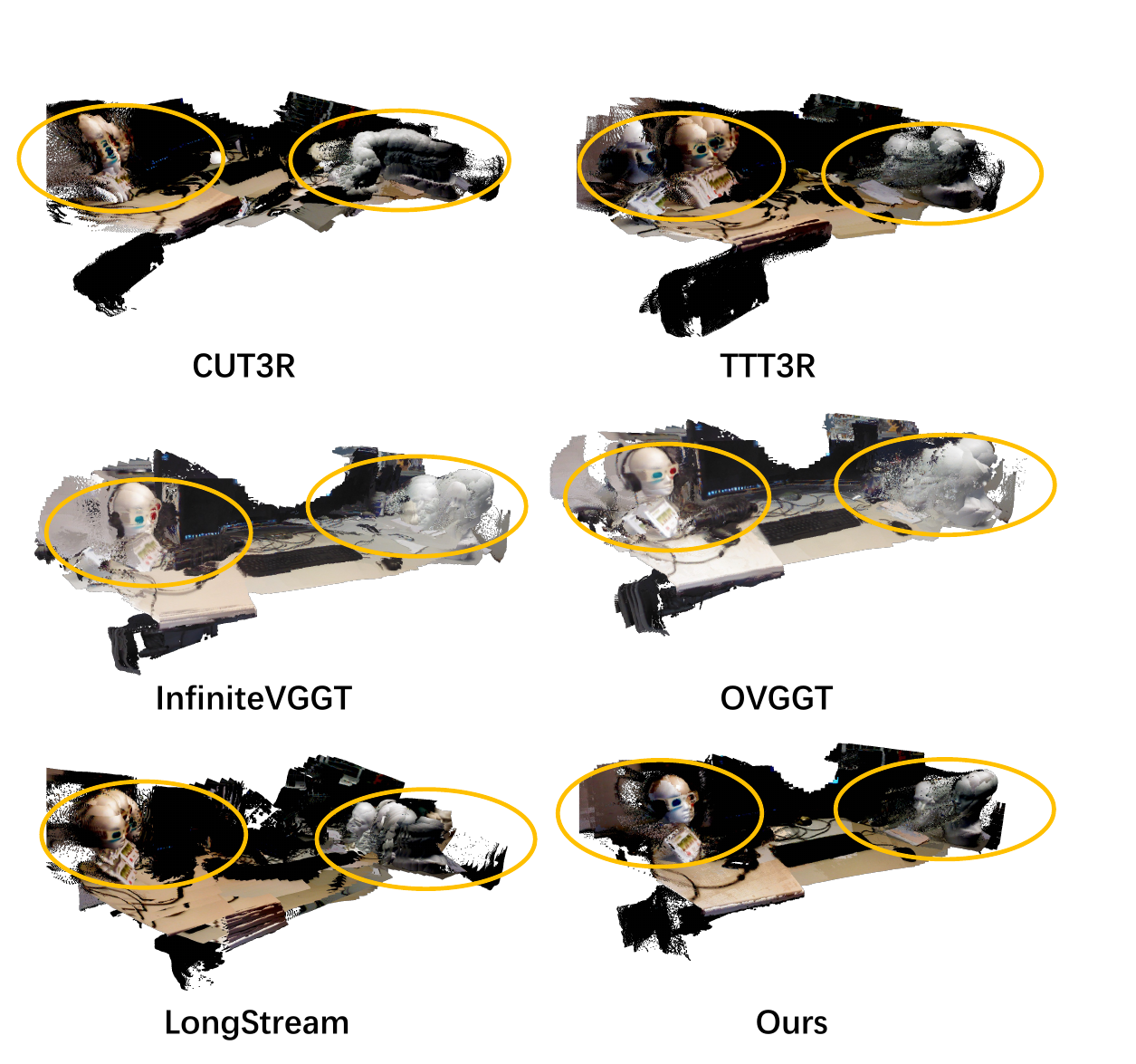}
\caption{
Qualitative Reconstruction Results on the 7-scenes dataset on long-sequence settings.
}
\label{fig:app5}
\end{figure}

\paragraph{Qualitative Results}
We provide additional results on the \textbf{mid-length} and \textbf{long-sequence}  setting, as shown in ~\cref{fig:app3} and \cref{fig:app5}. Our method produces more consistent geometric reconstruction results.

\section{Limitations and Future Work}
\label{sec.limitation}

While Stable3R achieves strong performance across multiple streaming reconstruction benchmarks, several directions remain for future work. First, our current study focuses on short-sequence training (up to 24 frames). Nevertheless, with a training-free extension strategy, our model generalizes well to 300-frame sequences and already achieves strong performance in long-horizon reconstruction. At the same time, concurrent works such as LongStream~\cite{cheng2026longstream}, LingBot-Map~\cite{chen2026geometric} and ABot-Recon~\cite{abot-recon} have shown that training on longer sequences can further benefit long-horizon reconstruction. Extending our framework to longer-sequence training is therefore a promising direction that we hope to explore as more computational resources become available.

Second, although our current framework is designed for stable streaming reconstruction, it does not explicitly model temporal correspondence over long horizons. Incorporating stronger temporal modeling and tracking mechanisms could further improve consistency across frames and broaden the applicability of the framework to more dynamic streaming scenarios.

Third, combining Stable3R's real-world reconstruction capabilities with recent generative models, such as ABot-Earth and GS-Voxel, could extend the framework beyond reconstruction toward more general 3D scene generation and world modeling~\cite{abot-earth-0.5, gs-voxel, sat3dgen, sat2density, sat2density++}.

Finally, an important future direction is to extend Stable3R toward dynamic-static decomposition, so that static scene geometry and moving objects can be reconstructed separately. We believe such a formulation would be especially valuable for dynamic environments and could serve as a step toward more general streaming 4D reconstruction.

\section{Societal Impacts}
\label{sec.social_impact}

\paragraph{Positive Social Impact.}

The advancement of our stable streaming 3D reconstruction technology can lead to several societal benefits: (1) Assistive Navigation: Reliable real-time 3D reconstruction can significantly enhance navigation aids for individuals with visual impairments, providing detailed spatial awareness of their surroundings. (2) Autonomous Systems: Improved stability in pose estimation and reconstruction is crucial for the safe operation of autonomous vehicles and embodied robots, potentially reducing accidents in complex environments. (3) Cultural Preservation: The technology can be utilized for the rapid and accurate digital archiving of historical sites and artifacts, preserving cultural heritage for future generations.

\paragraph{Potential Negative Impact.}
We also acknowledge the following risks associated with our technology: (1) Privacy Invasion: High-fidelity 3D reconstruction capabilities could potentially be misused for unauthorized surveillance. (2) Resource Consumption: Training and deploying deep learning models for 3D reconstruction require substantial computational resources, contributing to energy consumption. (3) Misuse in Deception: The reconstructed 3D models could theoretically be exploited to generate highly realistic synthetic data, which might be used to deceive other AI systems.

\end{document}